\documentclass{article}

\usepackage[a4paper, margin=1in]{geometry}
\usepackage{setspace}
\usepackage[nopatch=footnote]{microtype}
\usepackage{amsmath, amssymb, amsthm, bm}

\usepackage{graphicx}
\graphicspath{{./fig/}}
\usepackage{placeins}

\usepackage{authblk}

\usepackage{subcaption}
\usepackage{booktabs, threeparttable}

\usepackage[style=ieee, backend=biber]{biblatex}
\let\cite\autocite

\usepackage[title]{appendix}

\usepackage[colorlinks=true, linkcolor=black, citecolor=blue, urlcolor=blue]{hyperref}
\usepackage[capitalize]{cleveref}

\begin{document}

\title{ProMORNA: An End-to-End Generative Framework for Full-Length mRNA Design}

\renewcommand\Affilfont{\small\normalfont}

\author[3]{Zixi Shao}
\author[2]{Tao Wang}
\author[1]{Ye Shen}
\author[2]{Tianyi Huang\protect\footnotemark[1]}
\author[1]{Yibei Xiao\protect\footnotemark[1]}

\affil[1]{School of Pharmacy, China Pharmaceutical University}
\affil[2]{School of Artificial Intelligence, China Pharmaceutical University}
\affil[3]{Nanjing Foreign Language School}

\footnotetext[1]{Corresponding authors: Tianyi Huang (huangtianyi@cpu.edu.cn) and Yibei Xiao (yibei.xiao@cpu.edu.cn).}

\date{}

\maketitle

\begin{abstract}
Designing therapeutic messenger RNA (mRNA) requires generating full-length transcripts that optimally balance intracellular stability and translation efficiency. While generative models have advanced mRNA design, existing approaches rely on region-wise generation, expensive inference-time search, or target-specific model updates. To overcome these limitations, we introduce ProMORNA, a zero-shot, end-to-end generative framework for full-length mRNA design. ProMORNA aligns the design process with modern Large Language Model training pipelines: it utilizes a BART-style architecture pre-trained on over 6 million natural protein--mRNA pairs, followed by a two-stage post-training process comprising cold-start supervised fine-tuning and multi-objective reinforcement learning. By standardizing and aggregating relative advantages at the individual metric level with domain-specific stabilization, our post-training stage directly aligns the generative policy with conflicting biological constraints while avoiding direct scalarization of raw objective scores. \textit{In silico} evaluations demonstrate that ProMORNA pushes the Pareto frontier of predicted stability and translation efficiency beyond standard baselines and GEMORNA, a leading state-of-the-art method. In three independent \textit{in vitro} Fluc2P experiments, Candidate~2 achieved a 12--72\,h reporter-activity AUC 3.33 times that of the conventional Benchmark (Holm-adjusted $P=0.01285$), while its estimated 1.29-fold advantage over the reported best GEMORNA candidate remained exploratory. These results support ProMORNA as a promising framework for generating functionally active full-length mRNA candidates.
\end{abstract}

\section{Introduction}

The success of messenger RNA (mRNA) vaccines against SARS-CoV-2 has highlighted the broad potential of mRNA-based medicines~\cite{polack2020safety,baden2021efficacy,pardi2018mrna,qin2022mrna}. Creating effective mRNA medicines requires computational design of mRNA sequences based on the amino acid sequence of the desired protein~\cite{metkar2024tailor}.

Full-length mRNA design can be viewed as a conditional generation and multi-objective optimization problem: given a protein sequence of amino acids, the goal is to generate an mRNA sequence of nucleotides that, (i) translates back into that protein, and (ii) optimizes for certain competing biological objectives. Specifically, a full-length mRNA sequence has a central coding sequence (CDS) bordered by untranslated regions (UTRs) at both ends. The CDS is divided into codons of three adjacent nucleotides, where each codon naturally translates into a certain amino acid determined by the genetic code. Thus, the number of codons in the CDS must be equal to the number of amino acids in the target protein, and each codon must be selected from a fixed set of synonymous codons that translate into the amino acid in the corresponding position. In contrast, the UTRs are largely unconstrained but work with the CDS to shape biomolecular dynamics and finally protein output~\cite{sahin2014mrna, leppek2022combinatorial}.

\begin{figure}[ht]
  \centering
  \includegraphics[width=\linewidth]{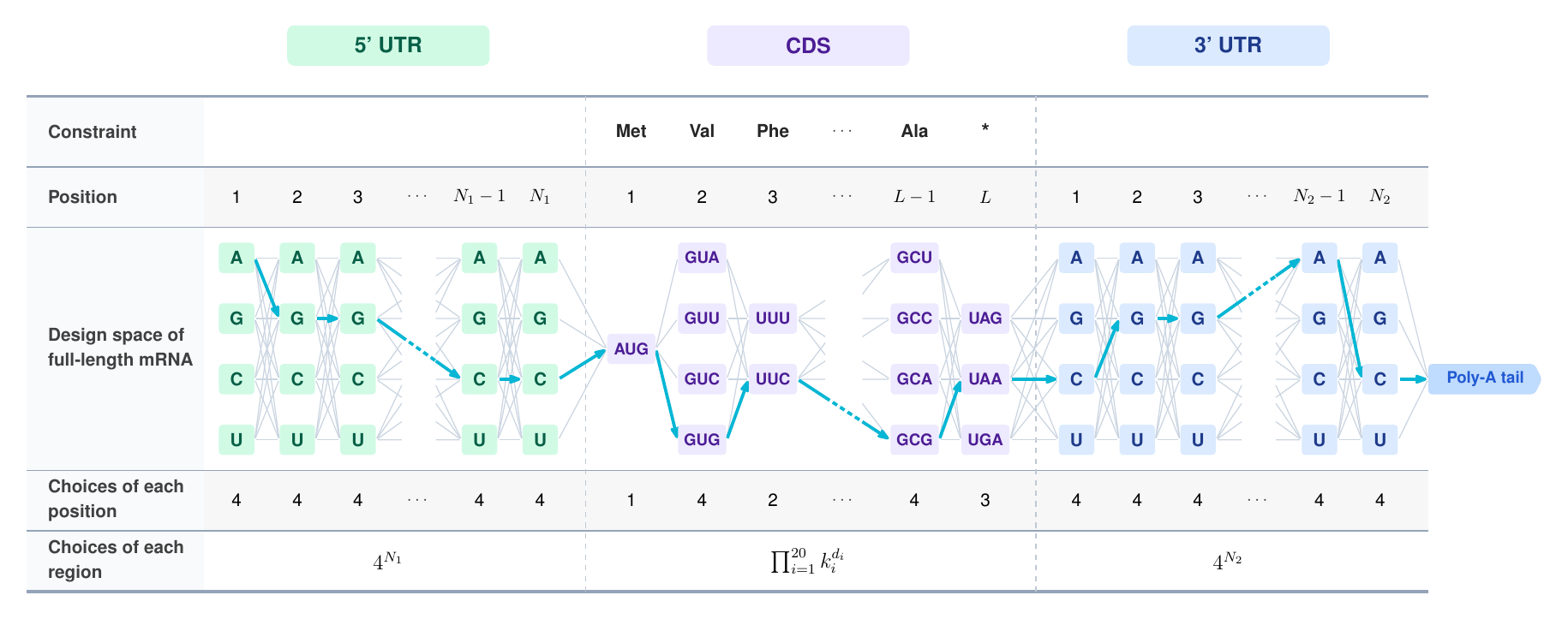}
  \caption{The vast potential mRNA search space of a target protein.}
  \label{fig:design-space}
\end{figure}

The primary challenge of mRNA design is the large design space, which makes exhaustive search infeasible. As illustrated in \cref{fig:design-space}, the number of possible CDSs alone for a given protein grows exponentially with the protein length, not even considering the unconstrained compositions of the UTRs.

A primary objective in mRNA sequence design is to maximize functional protein expression, which is shaped by stability and translation efficiency~\cite{mauger2019mrna,leppek2022combinatorial}. However, the complex biological mechanisms that affect these properties remain incompletely understood, making it difficult for rule-based algorithms to accurately predict and optimize for downstream biological properties. Conventional methods, such as LinearDesign~\cite{zhang2023algorithm}, approach this challenge by jointly optimizing simple heuristic proxy metrics like Minimum Free Energy (MFE) and Codon Adaptation Index (CAI)~\cite{sharp1987codon}. Although these algorithms can achieve mathematical optimality within their defined search spaces, their real-world biological performance is fundamentally bounded by the limitations of these rigid, rule-based heuristics.

Recently, generative models have emerged to address limitations of traditional heuristic algorithms. For example, mRNAutilus~\cite{patel2026mrnautilus} treated mRNA design as a discrete diffusion process guided by inference-time tree search; GEMORNA~\cite{zhang2025deep} used separate Transformer models to independently generate the 5' UTR, CDS, and 3' UTR. Most recently, mRNA-GPT~\cite{li2026mrna} proposed jointly optimizing these regions through full-length sequential generation using a unidirectional, decoder-only language model.

However, these existing generative frameworks are not strictly end-to-end and are limited by their architecture. mRNAutilus requires a CDS template and two UTR lengths as inputs to initialize the masked diffusion process, which is constrained by hand-crafted priors; GEMORNA does not directly optimize for downstream biological objectives, which limits the ability to explore candidates beyond the training distribution, and concatenates outputs from three independent models, which may overlook long-range context-dependency between different regions~\cite{leppek2022combinatorial,medjmedj2025evaluation}; mRNA-GPT is partially or fully blind to the protein target when generating the 5' UTR and CDS regions, limiting its capacity to plan beyond the current amino acid.

Furthermore, these methods rely on computationally intensive inference-time scaling: mRNAutilus runs tree search at every diffusion step; GEMORNA requires wet-lab screening on a large number of generated candidates and optionally utilizes beam search for CDS generation; mRNA-GPT applies Iterative Supervised Fine-Tuning (SFT) or Reinforcement Learning (RL) for each specific target from a fixed base model, generating thousands of candidates per iteration per target. The vast resources required for these models at inference time limit their application to practical scenarios such as personalized mRNA vaccines~\cite{sahin2018personalized, blass2021advances}, where rapid computational turnaround is critically required.

\begin{figure}[ht]
  \centering
  \includegraphics[width=\linewidth]{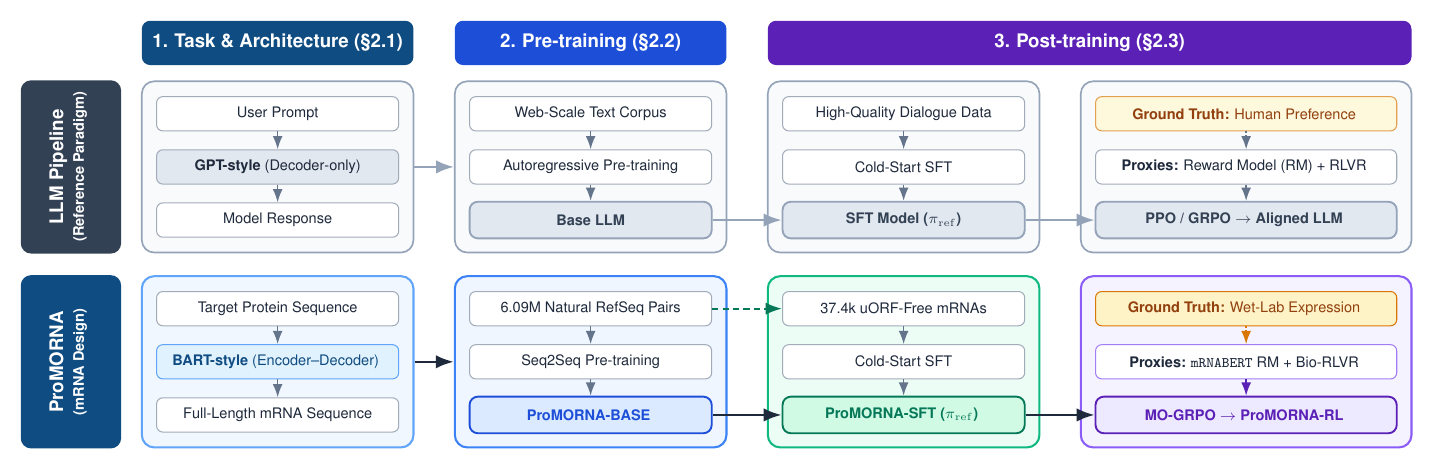}
  \caption{Overview of the ProMORNA framework and its conceptual alignment with modern LLM development.}
  \label{fig:pipeline-overview}
\end{figure}

To overcome these challenges, we propose \textbf{ProMORNA} (\textbf{Pro}tein-conditioned \textbf{M}ulti-\textbf{O}bjective \textbf{RNA} generation), an end-to-end generative framework for full-length mRNA design (\cref{fig:pipeline-overview}). First, we pre-train a BART-style encoder--decoder generative model on public data. Next, inspired by modern LLM alignment~\cite{instructgpt,deepseek-math,tulu3,deepseek-r1}, we employ a two-stage post-training pipeline consisting of cold-start supervised fine-tuning (SFT) followed by multi-objective reinforcement learning (RL) to optimize for competing biological metrics. \textit{In silico} experiments illustrate that ProMORNA improves the Pareto frontier for predicted stability and translation efficiency relative to standard supervised baselines and the existing generative mRNA design method, GEMORNA~\cite{zhang2025deep}. Wet-lab evaluation of seven Pareto-selected candidates on Fluc2P provides functional evidence for the generated designs.

We summarize the core contributions of our framework as follows:

\begin{itemize}

\item \textbf{Protein-conditioned end-to-end mRNA generation:} We introduce an encoder--decoder framework that generates complete mRNA transcripts, including the 5' UTR, CDS, and 3' UTR, directly from a target protein sequence, without requiring additional inputs at inference time. The model is given maximum freedom in generation and optimization, and only a minimal set of basic biological rules is enforced throughout the whole pipeline.

\item \textbf{Multi-stage, target-agnostic model alignment:} We structure full-length mRNA design into a progressive pre-training and two-stage post-training pipeline. Our target-agnostic RL post-training generalizes to unseen proteins, enabling rapid, high-quality zero-shot generation without requiring target-specific model updates. Furthermore, the autoregressive model natively supports inference-time scaling, allowing users to optionally generate and filter large candidate pools to further explore the Pareto frontier when computational budget permits.

\item \textbf{Pushing the Pareto frontier \textit{in silico}:} We evaluate ProMORNA \textit{in silico} against existing design strategies and assess generated candidates across translation efficiency, stability, normalized Minimum Free Energy (MFE), uridine content (U-content), and UTR length plausibility. Notably, we find that ProMORNA significantly improves the performance over supervised baselines and GEMORNA on the two primary predicted objectives: predicted stability and predicted translation efficiency.

\item \textbf{Prospective biological validation:} We synthesize and evaluate seven Pareto-selected ProMORNA candidates alongside a conventional mRNA Benchmark and the GEMORNA design GMR-FL-F5~\cite{zhang2025deep}. Three independent experiments identify candidates with higher time-integrated reporter activity than Benchmark.

\end{itemize}

\section{Methods}

In this section, we detail the \textbf{ProMORNA} framework illustrated in \cref{fig:pipeline-overview}. Following the paradigm of modern large language model (LLM) development, we organize our methodology into three parts: (i) task formulation and model architecture (\cref{sec:arch-data}); (ii) large-scale evolutionary pre-training (\cref{sec:pretraining}); and (iii) two-stage post-training (\cref{sec:posttraining}), which comprises a cold-start supervised fine-tuning (SFT) stage followed by multi-objective reinforcement learning (RL).

\subsection{Task Formulation and Model Architecture}
\label{sec:arch-data}

\paragraph{Task Formulation.} We formulate full-length mRNA design as a conditional sequence-to-sequence generation task. Given the amino acid sequence of a target protein $\mathbf{x}$, the goal is to generate a complete mRNA transcript $y = (y_1, \dots, y_T)$ spanning the 5' UTR, CDS, and 3' UTR by modeling the conditional probability distribution $P_\theta(y \mid \mathbf{x}) = \prod_{t=1}^{T} P_\theta(y_t \mid y_{<t}, \mathbf{x})$, such that the CDS translates back into $\mathbf{x}$ while the full transcript optimizes downstream biological properties.

\begin{figure}[ht]
  \centering
  \includegraphics[width=\linewidth]{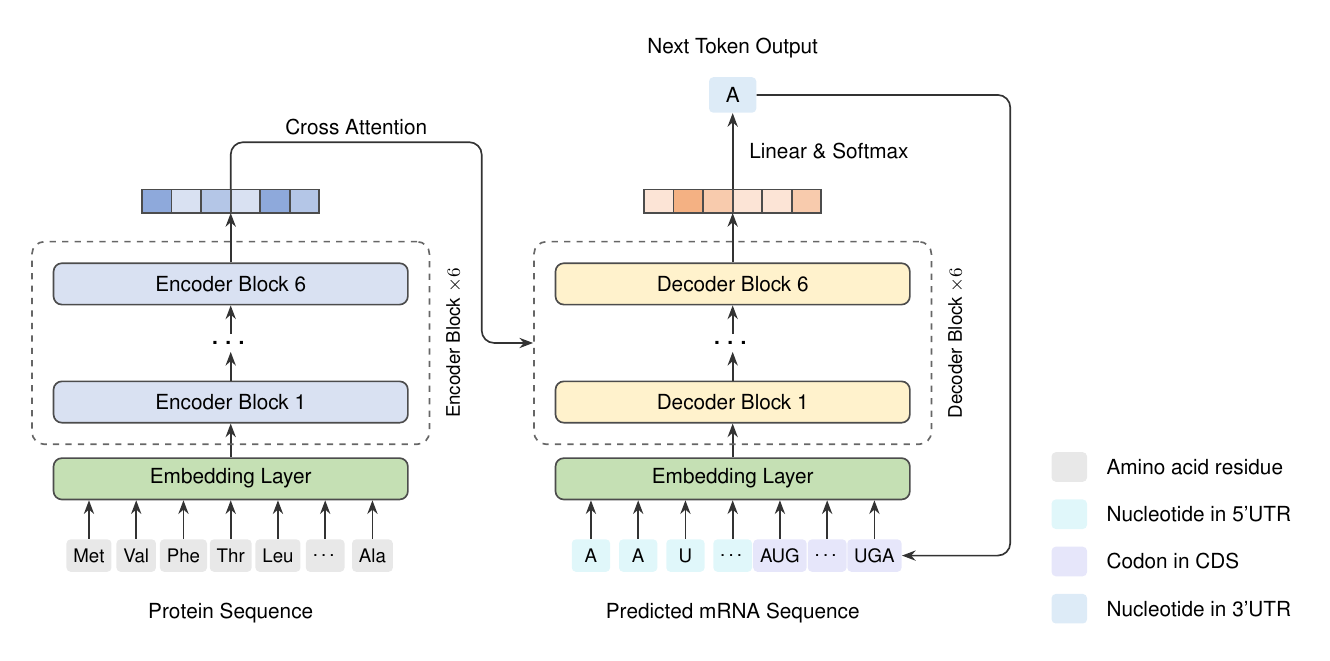}
  \caption{Modeling protein-to-mRNA with BART.}
  \label{fig:model-arch}
\end{figure}

\paragraph{Model Architecture.} Our model uses an encoder--decoder architecture based on the BART-small configuration~\cite{lewis2020bart} with a context limit of 1,024 tokens applied to both the encoder and the decoder, comprising 45.2M trainable parameters in total. As shown in \cref{fig:model-arch}, the encoder processes the input amino acid sequence of the target protein, while the decoder autoregressively generates the corresponding full-length mRNA sequence step by step. Because the encoder processes the full protein sequence before decoding begins, the decoder can condition every generated mRNA token on the complete target protein.

\paragraph{Tokenization.} We apply per-amino-acid tokenization for the source protein sequences. For target mRNA sequences, we use a hybrid vocabulary: per-nucleotide tokenization for both UTRs, and per-codon tokenization for the CDS. Special separator tokens are inserted to mark region boundaries and prompt the model to switch between nucleotide-level and codon-level vocabularies.

\subsection{Pre-training}
\label{sec:pretraining}

\paragraph{Training Dataset.} We curated vertebrate transcripts from NCBI RefSeq~\cite{oleary2016refseq}, prioritizing shorter sequences to ensure computational tractability and to remain within a practical length regime for future \textit{in vitro} synthesis. We filtered out sequences labeled as ``pseudo'' and retained transcripts with canonical start and stop codons, a verified N-terminal methionine, and minimum 5' and 3' UTR lengths of 20\,nt and 30\,nt, respectively. Additionally, all retained sequences were required to fit within our 1,024-token limit under our hybrid tokenization scheme. This pipeline yielded 6,089,519 natural protein--mRNA pairs. From these data, the model learns the conditional distribution $P(y \mid \mathbf{x})$ of a full-length mRNA transcript $y$ given a target protein sequence $\mathbf{x}$, implicitly capturing natural vertebrate codon usage biases and plausible UTR compositions.

\paragraph{Training Objective.} The model is pre-trained by minimizing the standard sequence-to-sequence negative log-likelihood:
\begin{equation}
\mathcal{L}(\theta) = - \frac{1}{T} \sum_{t=1}^{T} \log P_\theta(y_t \mid y_{<t}, \mathbf{x}),
\label{eq:nll}
\end{equation}
where $\theta$ represents the trainable parameters of the model, $T$ is the total length of the target mRNA sequence, $y_t$ is the ground-truth target token at time step $t$, $y_{<t}$ denotes the sequence of preceding target tokens, and $\mathbf{x}$ is the tokenized source input amino acid sequence. We denote the resulting pre-trained checkpoint as \textbf{ProMORNA-BASE} (or \textbf{BASE}).

\subsection{Post-training}
\label{sec:posttraining}

While large-scale pre-training captures natural evolutionary sequence grammar, designing therapeutic mRNAs requires shifting the model's generative prior toward high protein expression and explicitly optimizing competing biochemical properties. We therefore employ a two-stage post-training pipeline consisting of a cold-start supervised fine-tuning (SFT) stage (\cref{sec:sft}) followed by online reinforcement learning (\cref{sec:metrics,sec:mo-grpo}).

\subsubsection{Cold-Start Supervised Fine-Tuning}
\label{sec:sft}

Aligning with modern LLM post-training paradigms~\cite{instructgpt,deepseek-r1}, we employ SFT to establish a high-quality cold start for reinforcement learning. While pre-training captures broad natural evolutionary distributions, therapeutic mRNAs require maximized protein expression. Therefore, we curate a purified subset of 37,369 NM-prefixed RefSeq mRNAs strictly lacking upstream open reading frames (uORFs), which are natural regulatory elements that suppress translation~\cite{calvo2009upstream}. Fine-tuning the base model on this subset using the objective in \cref{eq:nll} shifts its generative prior toward therapeutic efficacy, yielding the \textbf{ProMORNA-SFT} (or \textbf{SFT}) checkpoint. Crucially, this step provides a biologically aligned reference policy $\pi_{\text{ref}}$ for the subsequent RL phase, ensuring that the KL-divergence penalty does not force the model to mimic undesirable natural traits.

\subsubsection{Biological Objectives and Design Constraints}
\label{sec:metrics}

To guide policy optimization toward high-performing therapeutic candidates, we define $K=5$ biological objectives that jointly capture functional efficacy, structural thermodynamics, sequence composition, and length regularity. For the $i$-th generated mRNA sequence, let $r_{i,k}$ denote its raw score on the $k$-th objective:

\paragraph{Predicted Stability ($k=1$).} Intracellular mRNA stability governs resistance to enzymatic degradation and determines the functional half-life of the transcript~\cite{mauger2019mrna,leppek2022combinatorial}. Because stability cannot be computed analytically from primary sequence alone, we estimate $r_{i,1}$ using a ridge-regression head $f_\mathrm{S}$ trained on global-average-pooled sequence embeddings from a frozen mRNABERT encoder~\cite{xiong2025mrnabert} (\cref{fig:mrnabert}; see \cref{appendix:model-training-setup} for training details). This primary functional objective is maximized.

\begin{figure}[ht]
  \centering
  \includegraphics[width=\linewidth]{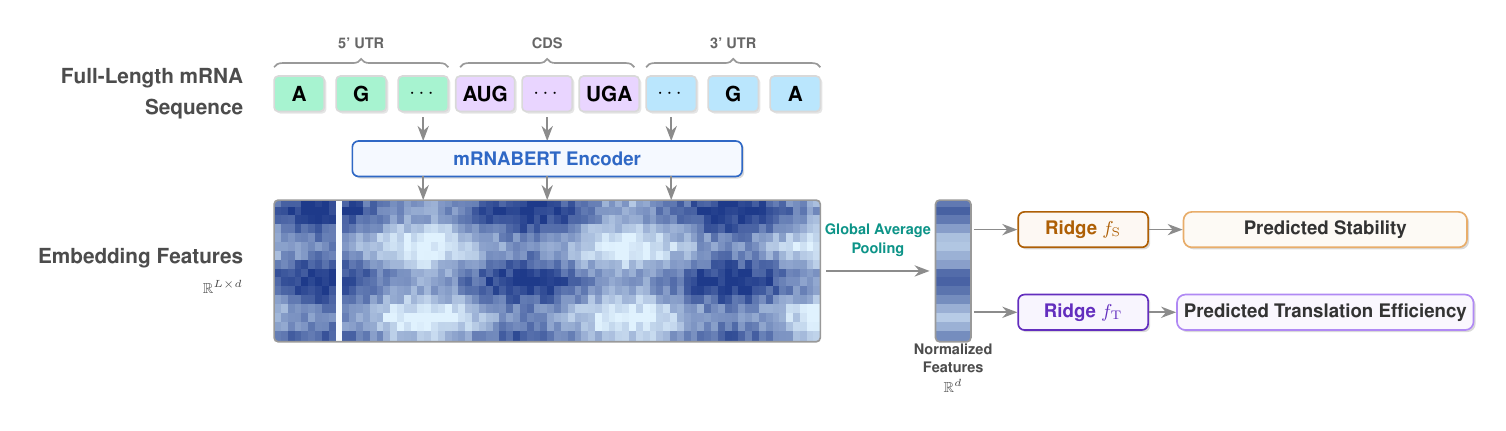}
  \caption{Predicting mRNA stability and translation efficiency with mRNABERT.}
  \label{fig:mrnabert}
\end{figure}

\paragraph{Predicted Translation Efficiency ($k=2$).} Translation efficiency measures the rate at which ribosomes initiate and elongate along the transcript to synthesize the target protein. Analogous to stability, we predict $r_{i,2}$ using an independent ridge-regression head $f_\mathrm{T}$ operating on the same pooled mRNABERT representation (\cref{fig:mrnabert}). This second primary functional objective is also maximized.

\paragraph{Normalized Minimum Free Energy (MFE, $k=3$).} Secondary structure folding critically modulates both physical transcript longevity and ribosomal transit~\cite{mauger2019mrna,zhang2023algorithm}. We compute the full-transcript folding free energy using LinearFold~\cite{huang2019linearfold} and normalize it by the total nucleotide length to obtain a length-invariant proxy for global thermodynamic stability. Since more stable secondary structures correspond to more negative free energy values, this objective is minimized.

\paragraph{U-content ($k=4$).} U-content is defined as the fractional proportion of uridine nucleotides across the full-length transcript. Because AU-rich elements accelerate mRNA decay~\cite{barreau2005aurich} and uridine-rich motifs trigger innate immune receptors that suppress translation~\cite{kariko2005suppression,verbeke2022innate}, we explicitly minimize global uridine fraction.

\paragraph{UTR Length Plausibility ($k=5$).} To prevent length degeneration during unconstrained UTR generation, we regularize the generated 5' and 3' UTR lengths using Gaussian-shaped priors derived from natural transcripts, maximizing:
\begin{equation}
r_{i,5} =
\sum_{j \in \{5,3\}}
\exp\left(-\frac{1}{2}\left(\frac{u_{i,j}-\mu_{i,j}}{\sigma_j}\right)^2\right),
\end{equation}
where $u_{i,j}$ denotes the nucleotide length of the generated $j$' UTR for sequence $i$, $\mu_{i,j}$ is the corresponding wild-type UTR length of the prompt protein in the training dataset, and $\sigma_j$ is the global standard deviation of $j$' UTR lengths across the dataset. This term serves solely as a training-time regularizer, requiring no wild-type length priors during zero-shot generation or evaluation.

\paragraph{Generation Constraints.} To guarantee sequence validity throughout rollouts, we combine prefix-level constrained decoding with a transcript-level post-hoc validity check. During autoregressive decoding, token sampling within the CDS is masked to allow only synonymous codons of the corresponding target amino acid, whereas UTR decoding is restricted to standard nucleotide tokens and region separators. Because prefix constraints alone cannot prevent premature truncation at the 1,024-token context limit or enforce global structural rules, each completed rollout undergoes a post-hoc verification of region boundaries, minimum UTR lengths, canonical start and stop codons, reading-frame integrity, and exact translation into the target protein. Moreover, following LinearDesign~\cite{zhang2023algorithm}, transcripts containing contiguous double-stranded regions of 33 or more base pairs are rejected to mitigate double-stranded RNA-mediated innate immune responses. Sequences failing any post-hoc check bypass predictor evaluation, are excluded from group mean and standard deviation calculations, and receive a deterministic maximal penalty during advantage assignment.

\subsubsection{Multi-Objective Reinforcement Learning}
\label{sec:mo-grpo}

While standard RL algorithms collapse multiple evaluation metrics into a single explicit scalar reward, this static scalarization can be unstable for biological sequence design: biological metrics have different scales, and their variances shift asynchronously as the policy evolves. Recent studies in multi-objective LLM alignment, specifically Multi-Objective GRPO (MO-GRPO)~\cite{ichihara2026mogrpo} and Group reward-Decoupled Normalization Policy Optimization (GDPO)~\cite{liu2026gdpo}, have shown that applying vanilla GRPO~\cite{deepseek-math} to summed rewards biases optimization toward high-variance objectives and causes reward signal collapse. Both methods address this by standardizing each objective independently within a rollout group prior to aggregation.

\begin{figure}[ht]
  \centering
  \includegraphics[width=\linewidth]{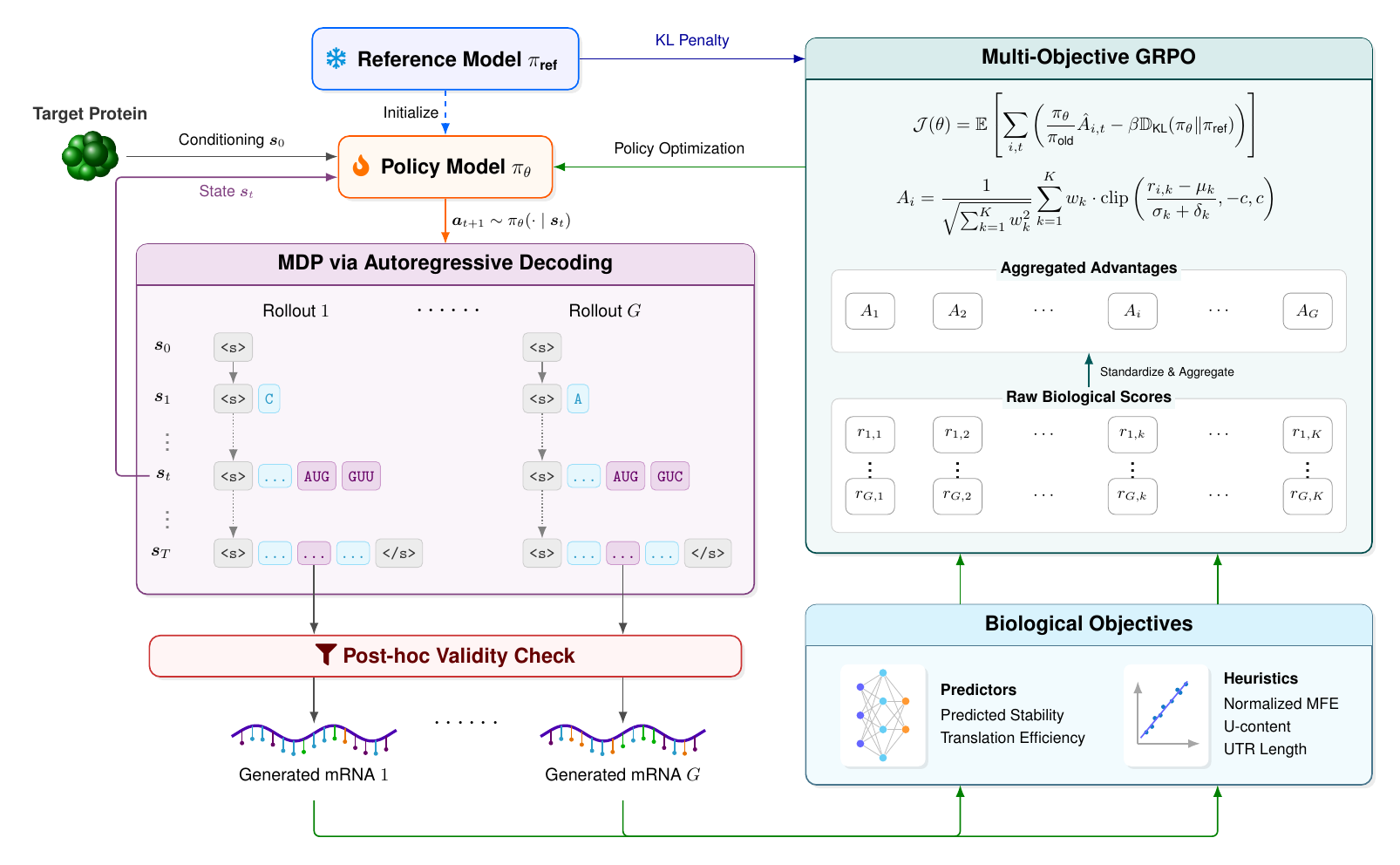}
  \caption{ProMORNA's post-training framework. \texttt{<s>} is the start token and \texttt{</s>} is the end token. Note that this is only a demonstration of the rollout process; in practice, mRNA sequences of varying lengths can be generated for the same target protein.}
  \label{fig:GRPO}
\end{figure}

In ProMORNA, we adopt this decoupled group-relative standardization framework (retaining the name MO-GRPO, \cref{fig:GRPO}) and introduce several domain-specific improvements to address practical instabilities in biological sequence optimization.

\paragraph{$L_2$-Normalized Weighting and Dynamic Variance Damping.}
For a given prompt, the policy samples a group of $G$ complete mRNA sequences. Let $\mu_k$ and $\sigma_k$ be the empirical mean and standard deviation of the $k$-th objective scores across valid sequences within the current group. Extending the unweighted per-objective standardization in MO-GRPO~\cite{ichihara2026mogrpo}, the aggregated baseline advantage $A_i$ for sequence $i$ is formulated as:
\begin{equation}
A_i = \frac{1}{\sqrt{\sum_{k=1}^K w_k^2}} \sum_{k=1}^K w_k \cdot \operatorname{clip}\left( \frac{r_{i,k} - \mu_k}{\sigma_k + \delta_k}, -c, c \right),
\label{eq:mogrpo-adv}
\end{equation}
where $w_k$ denotes the relative importance and optimization direction of the $k$-th objective ($w_k > 0$ for maximization and $w_k < 0$ for minimization). Normalizing the weights by their $L_2$ norm mathematically guarantees that, assuming the objectives are mutually orthogonal, the standard deviation of the aggregated advantages $\mathrm{std}(\{A_1,\dots,A_G\})$ remains approximately $1$. Crucially, whereas prior MO-GRPO~\cite{ichihara2026mogrpo} divides directly by $\sigma_k$, we introduce a dynamic damping factor $\delta_k > 0$ for each objective. When the group variance $\sigma_k$ drops significantly ($\sigma_k \ll \delta_k$), $\delta_k$ smoothly pushes the normalized advantage contribution of this metric toward zero. This naturally prevents numerical instability and shifts the model's exploration focus toward other under-optimized metrics.

\paragraph{Component-Wise Extreme-Value Clipping.}
Unlike simple, rule-based NLP rewards, biological reward components are computed from heuristic or learned proxy predictors (\texttt{mRNABERT} and \texttt{LinearFold}), whose outputs on out-of-distribution rollouts may occasionally contain large deviations. In contrast to prior formulations~\cite{ichihara2026mogrpo,liu2026gdpo}, we apply component-wise extreme-value clipping with bound $c=3.0$ (\cref{eq:mogrpo-adv}) prior to aggregation to limit the influence of outlier objective values.

\paragraph{Mean-Only Post-Centralization and Validity-Masked Penalty Injection.}
Because clipping, damping, and invalid-sequence penalties can shift the group-level mean away from zero, we apply a post-centralization step to obtain the centered sequence-level advantage $\hat{A}_i$, which is then broadcast across all tokens within the $i$-th sequence to serve as the token-level advantage $\hat{A}_{i,t}$:
\begin{equation}
\hat{A}_{i,t} = \hat{A}_i = A_i - \frac{1}{G}\sum_{j=1}^G A_j, \quad \forall t \in \{1, \dots, |\bm{a}_i|\}.
\label{eq:post-center}
\end{equation}
Importantly, unlike GDPO~\cite{liu2026gdpo}, which re-divides the aggregated advantage sum by the batch standard deviation, we do not re-divide by the standard deviation in \cref{eq:post-center}. This preserves the adaptive magnitude derived from the damping factor $\delta_k$, allowing the advantage to decay naturally as the policy converges. For invalid sequences, we exclude them from $\mu_k$ and $\sigma_k$ calculation and directly assign a maximal penalty $A_i = -c\sum |w_k|/\sqrt{\sum w_k^2}$ before post-centralization, thereby incorporating invalid samples into the policy update.

\paragraph{Token-Level Policy Gradient and Unbiased KL Regularization.}
Finally, the policy $\pi_\theta$ is updated to maximize:
\begin{equation}
\mathcal{J}(\theta) = \mathbb{E}_{\substack{\bm{s}_0 \sim P(S_0) \\ \{\bm{a}_i\}_{i=1}^G \sim \pi_{\text{old}}(\cdot|\bm{s}_0)}} \left[ \frac{1}{\sum_{i=1}^G|\bm{a}_i|} \sum_{i=1}^{G} \sum_{t=1}^{|\bm{a}_i|} \left( \frac{\pi_{\theta}(\bm{a}_{i,t}|\bm{s}_{i,t})}{\pi_{\text{old}}(\bm{a}_{i,t}|\bm{s}_{i,t})} \hat{A}_{i,t} - \beta \, \mathbb{D}_{\text{KL}}\Big[ \pi_{\theta} \| \pi_{\text{ref}} \Big] \right) \right],
\end{equation}
where the KL divergence is approximated analytically as
\begin{equation}
\mathbb{D}_{\text{KL}}(\pi_\theta \| \pi_{\text{ref}}) = \frac{\pi_{\theta}(\bm{a}_{i,t}|\bm{s}_{i,t})}{\pi_{\text{old}}(\bm{a}_{i,t}|\bm{s}_{i,t})} \left( \frac{\pi_{\text{ref}}(\bm{a}_{i,t}|\bm{s}_{i,t})}{\pi_{\theta}(\bm{a}_{i,t}|\bm{s}_{i,t})} - \log \frac{\pi_{\text{ref}}(\bm{a}_{i,t}|\bm{s}_{i,t})}{\pi_{\theta}(\bm{a}_{i,t}|\bm{s}_{i,t})} - 1 \right).
\end{equation}
Here, different from vanilla GRPO~\cite{deepseek-math}, we (i) use a token-level policy gradient loss~\cite{yu2025dapo}, assigning equal weights to every generated token; (ii) estimate the KL penalty with an importance-weighted estimator~\cite{liu2025deepseek}, which eliminates bias introduced by evaluating samples drawn from the old policy.

\section{Results}

This section presents \textit{in silico} and \textit{in vitro} results of ProMORNA on the widely used Fluc2P target. We compare ProMORNA with GEMORNA and supervised baselines to evaluate its mRNA generation performance. \Cref{appendix:exp-setup} presents the experimental setup, including the target protein, baselines, parameters, generation protocol, and the reward predictor training. \Cref{appendix:rl-dynamics} presents RL learning curves. Ablation studies demonstrate that standard scalarized reward optimization suffers from rapid reward hacking and KL-collapse, underscoring the necessity of our metric-level advantage formulation during RL post-training (\cref{appendix:mo-grpo-ablation}). Our additional analyses (see \cref{appendix:additional-analysis}) demonstrate that ProMORNA automatically learned several known biological rules, such as optimized GC content and structures more conducive to regulating protein expression. These findings may further indicate that the model enhanced its understanding of sequence structure in the absence of explicit supervisory signals.

\subsection{\textit{In silico} mRNA Generation Performance}

In this subsection, we evaluate ProMORNA on Fluc2P using the reward-related metrics. We compare ProMORNA with GEMORNA and the two preceding checkpoints from our training pipeline, BASE and SFT.

\begin{figure}[ht]
  \centering

  \begin{subfigure}[b]{0.33\linewidth}
    \centering
    \includegraphics[width=0.91\linewidth]{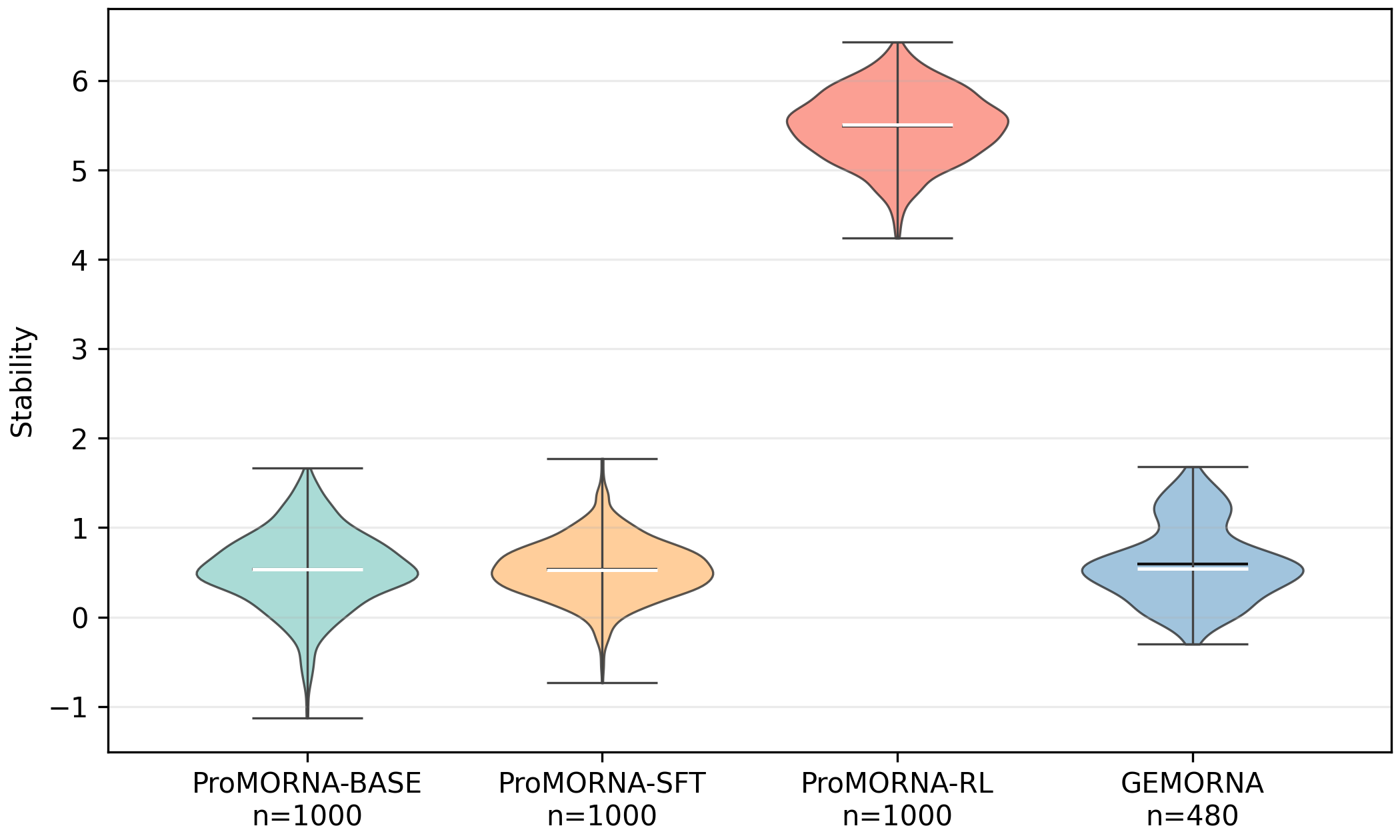}
    \caption{Predicted Stability}
    \label{fig:stability}
  \end{subfigure}%
  \begin{subfigure}[b]{0.33\linewidth}
    \centering
    \includegraphics[width=0.91\linewidth]{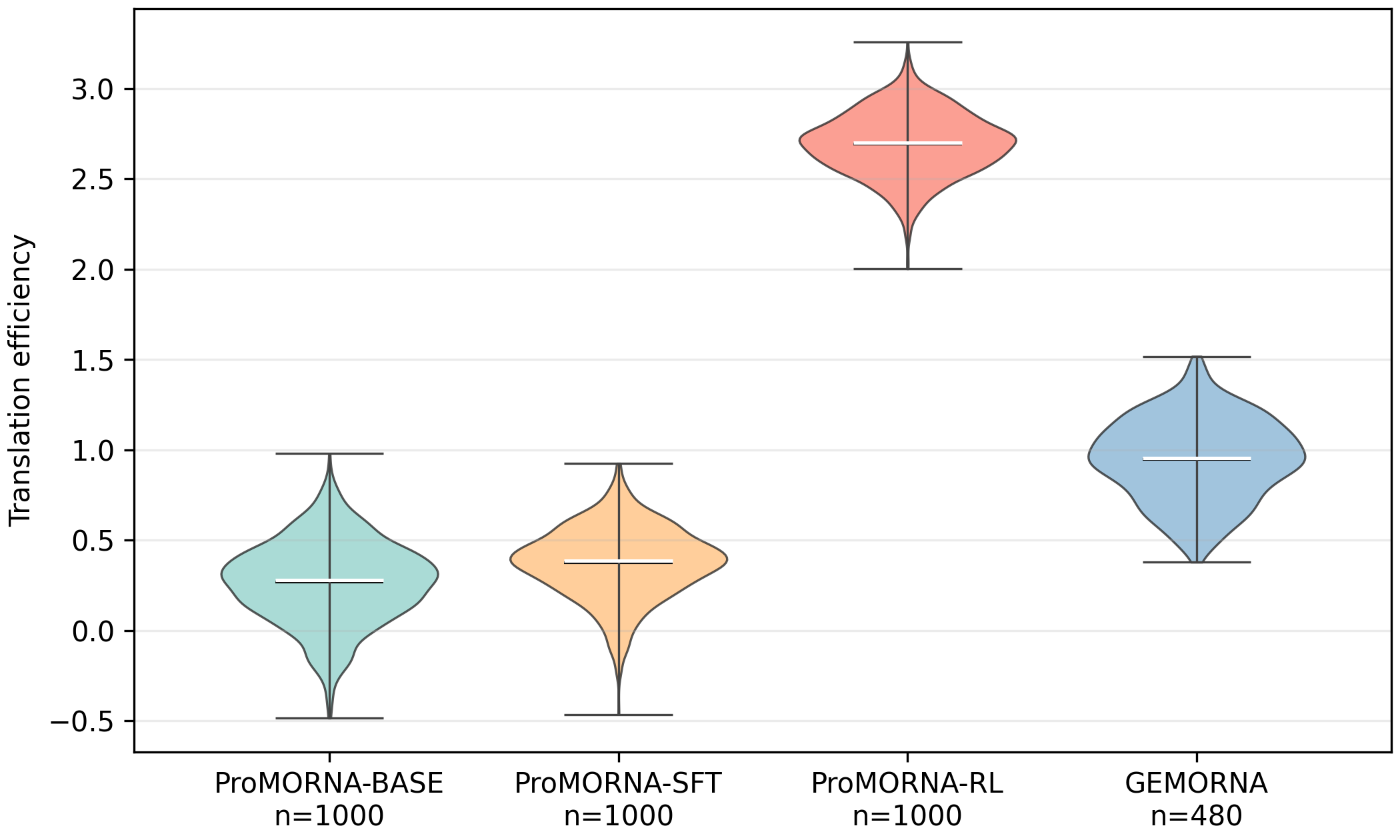}
    \caption{Predicted Translation Efficiency}
    \label{fig:te}
  \end{subfigure}%
  \begin{subfigure}[b]{0.33\linewidth}
    \centering
    \includegraphics[width=0.91\linewidth]{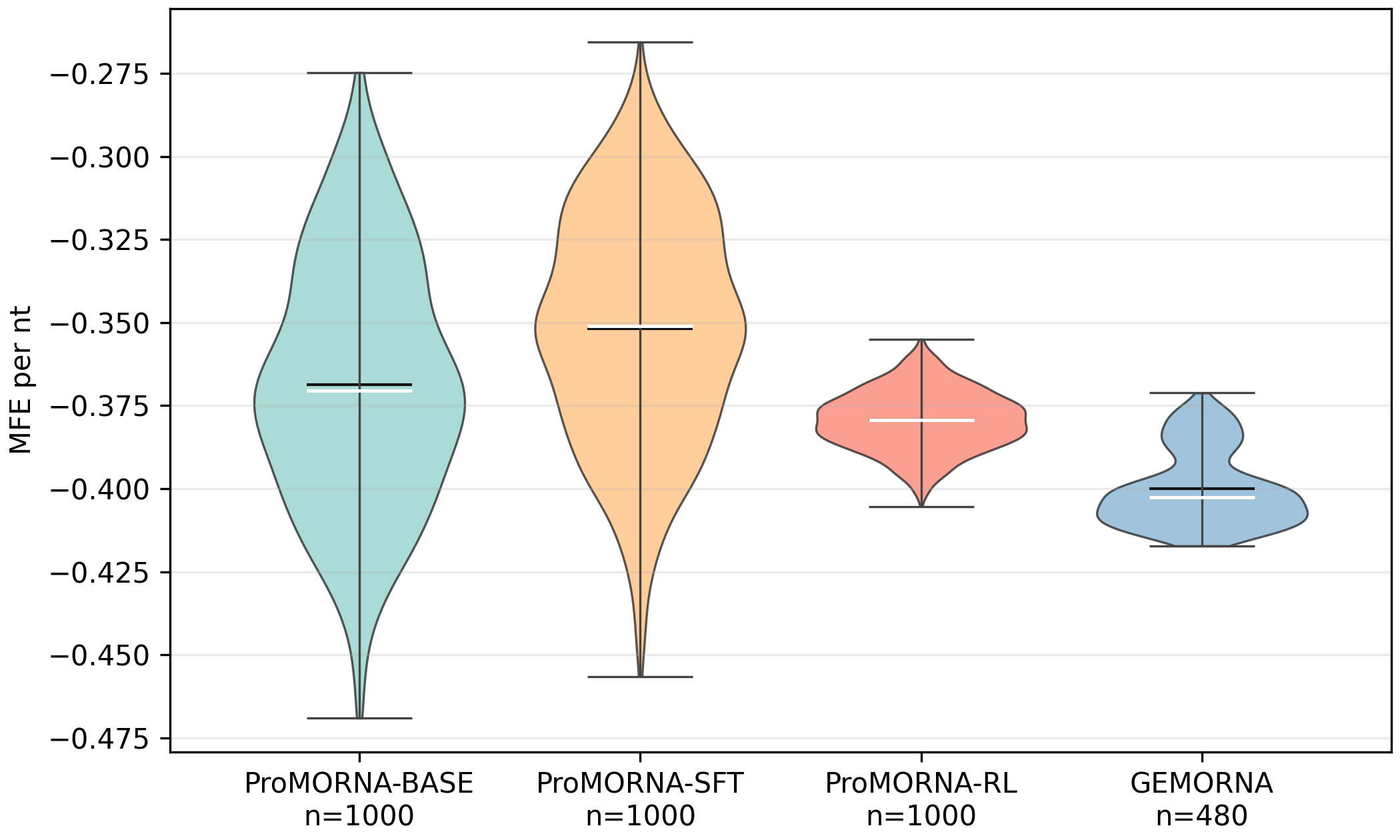}
    \caption{Normalized MFE}
    \label{fig:mfe}
  \end{subfigure}

  \vspace{0.8em}

  \begin{subfigure}[b]{0.33\linewidth}
    \centering
    \includegraphics[width=0.91\linewidth]{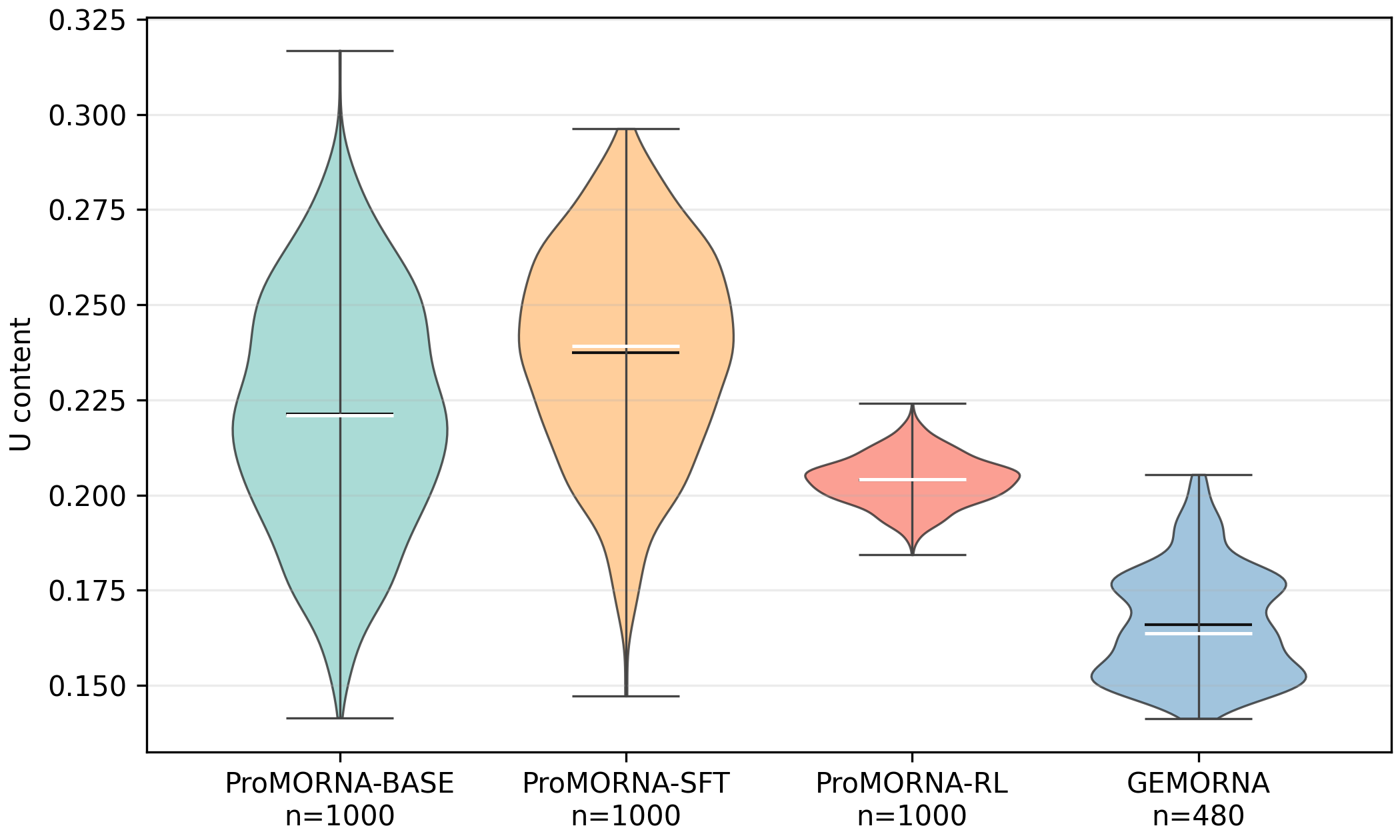}
    \caption{U-content}
    \label{fig:ucontent}
  \end{subfigure}%
  \begin{subfigure}[b]{0.33\linewidth}
    \centering
    \includegraphics[width=0.91\linewidth]{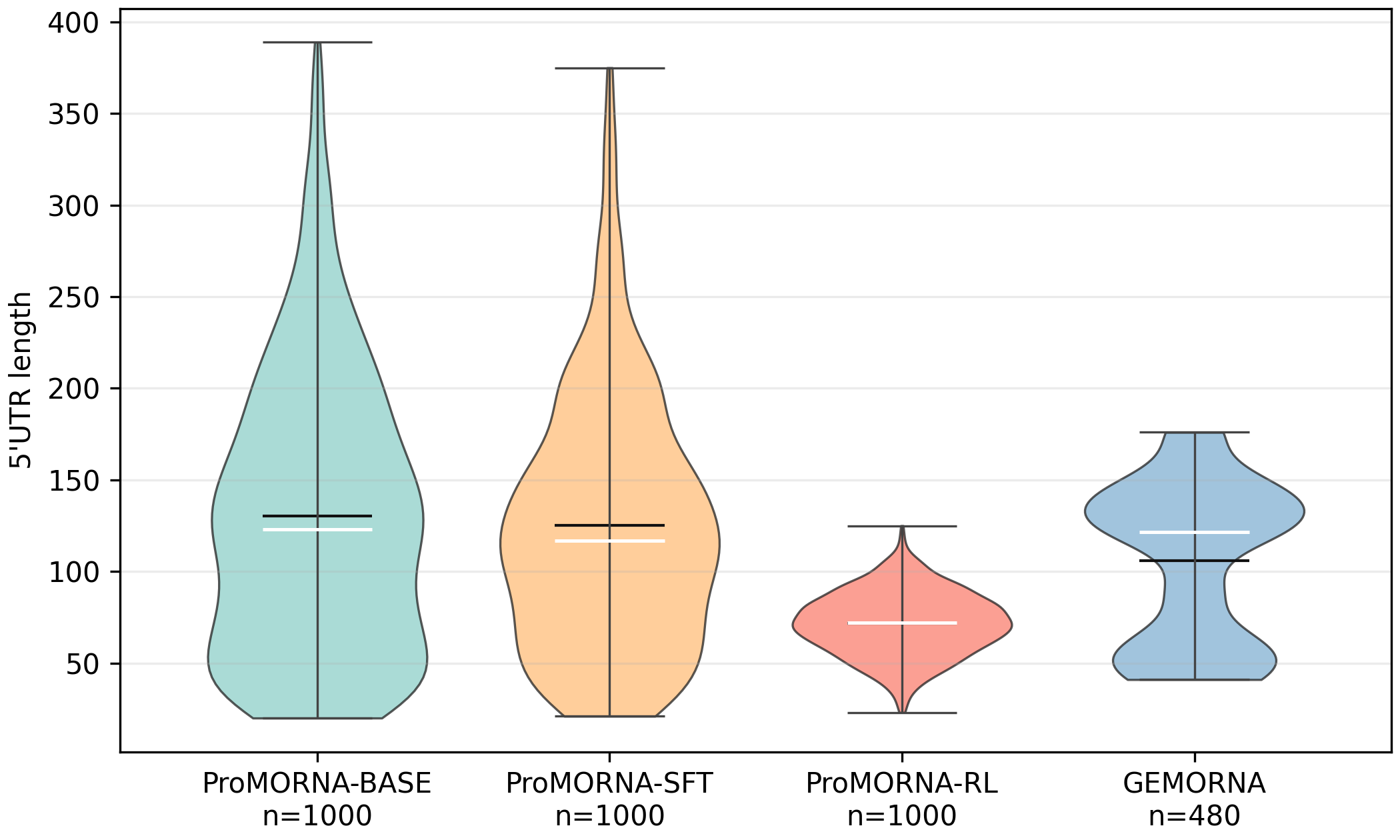}
    \caption{5' UTR length}
    \label{fig:5ulen}
  \end{subfigure}%
  \begin{subfigure}[b]{0.33\linewidth}
    \centering
    \includegraphics[width=0.91\linewidth]{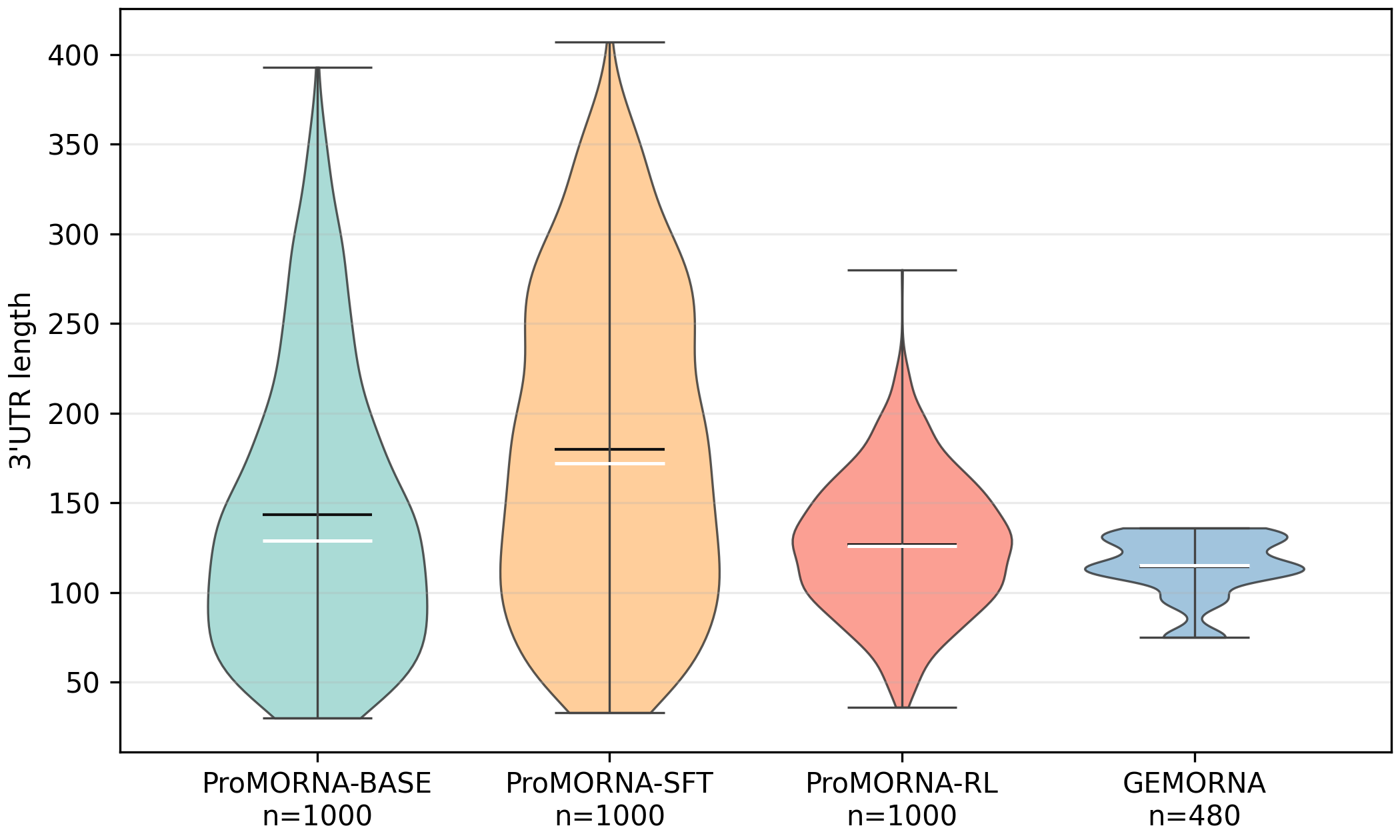}
    \caption{3' UTR length}
    \label{fig:3ulen}
  \end{subfigure}

  \caption{Comparison of three ProMORNA checkpoints and GEMORNA on reward-related metrics.}
  \label{fig:ec}
\end{figure}

The comparison results and ablation study are presented in \cref{fig:ec}. As shown in \cref{fig:stability,fig:te}, ProMORNA scores highest on the two primary functional objectives: predicted stability and translation efficiency. This indicates that our RL post-training effectively guides the generator toward candidates with improved functional proxy scores. While GEMORNA also outperforms the purely supervised BASE and SFT models, consistent with its strong region-wise mRNA generation design, ProMORNA builds on this by directly optimizing the full-length transcript using multi-objective reinforcement learning, rather than recombining independently generated UTR and CDS elements.

\Cref{fig:mfe,fig:ucontent} show that ProMORNA scores slightly higher than GEMORNA scores on normalized full-length MFE and U-content, but lower than BASE and SFT scores. Because mRNA expression relies on complex interactions among regional structure, codon usage, UTR context, innate immune sensing, and degradation, these secondary metrics should be evaluated alongside the primary functional objectives~\cite{metkar2024tailor}. ProMORNA's moderate scores on these subsidiary metrics reflect the core goal of our multi-objective post-training: prioritizing a balanced, full-length trade-off over maximizing individual proxy metrics independently.

Regarding UTR length distributions (\cref{fig:5ulen,fig:3ulen}), ProMORNA tends to generate shorter 5' UTRs and longer 3' UTRs compared to the supervised baselines. Since UTR length effects are highly context-dependent~\cite{medjmedj2025evaluation}, this distribution suggests that the length-plausibility reward and functional objectives work together to guide ProMORNA toward an optimal, target-specific UTR length for Fluc2P instead of defaulting to a universally fixed length.

\begin{figure}[ht]
  \centering
  \includegraphics[width=0.8\linewidth]{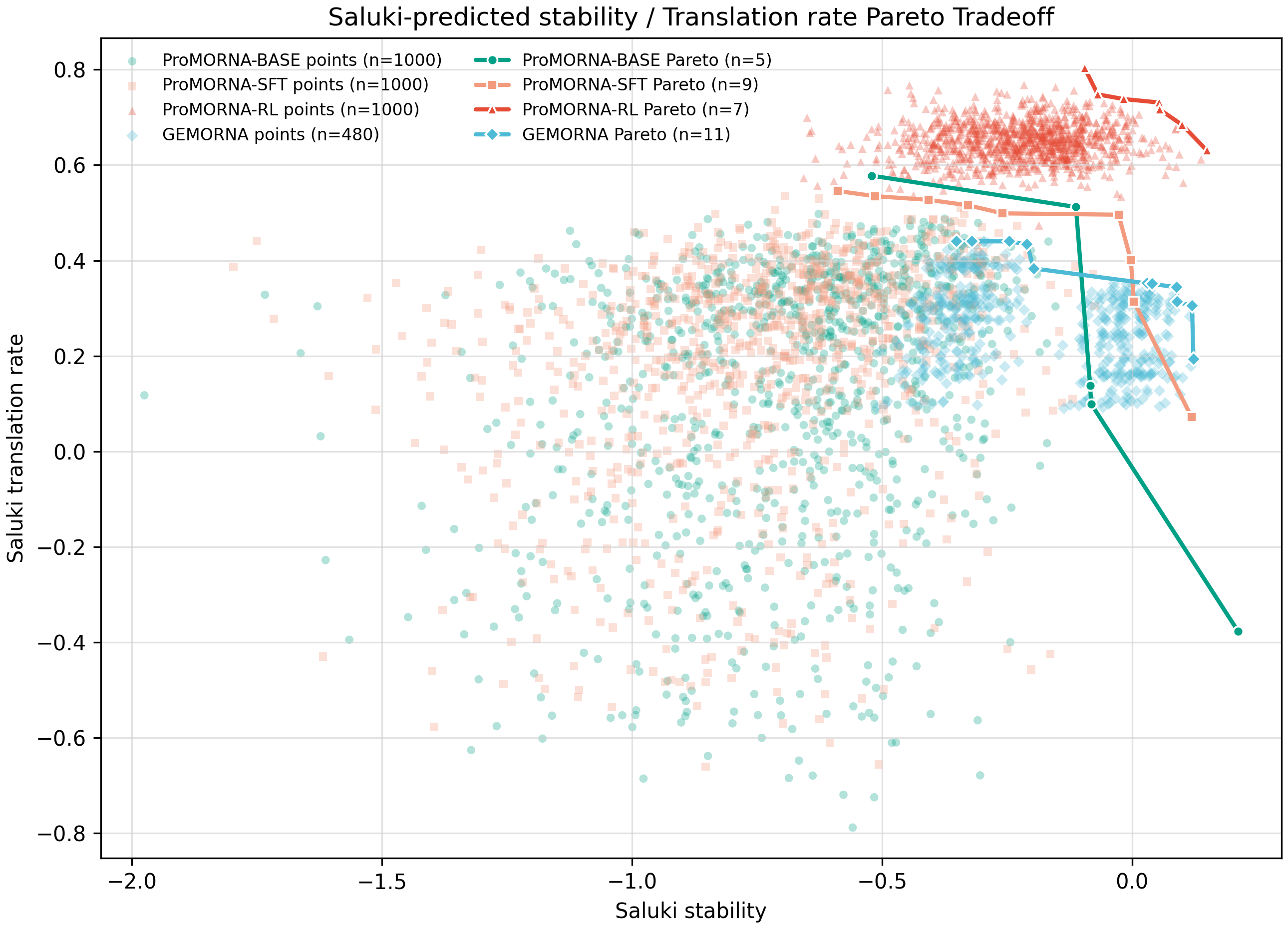}
  \caption{Orthogonal evaluation of Pareto frontiers for predicted stability and translation rate across four models using the independent evaluator Saluki~\cite{agarwal2022genetic}.}
  \label{fig:saluki-pareto}
\end{figure}

Because the stability and translation efficiency scores in \cref{fig:stability,fig:te} are derived from the same mRNABERT-based proxy predictors used during RL training, optimizing directly against them could raise concerns of proxy overfitting. To provide an independent, orthogonal evaluation, we further assess all generated candidates using Saluki~\cite{agarwal2022genetic}, an external sequence-based deep learning model that was entirely withheld from our training pipeline. \Cref{fig:saluki-pareto} illustrates the resulting Saluki-predicted stability and translation rate distributions alongside their corresponding Pareto frontiers across three ProMORNA models and GEMORNA. Consistent with our internal metrics, ProMORNA-RL achieves a pronounced upper-right shift of the Pareto frontier relative to both supervised baselines and GEMORNA. This cross-evaluator consistency confirms that our multi-objective RL post-training captures genuine sequence-level determinants of stability and translation rather than exploiting artifacts of the training reward predictors, providing a reliable basis for candidate prioritization in subsequent wet-lab validation.

\subsection{\textit{In Vitro} Validation}

Building on the orthogonal \textit{in silico} evaluation, we select all seven Pareto-optimal candidates identified by Saluki from the 1,000 ProMORNA-RL sequences (\cref{fig:saluki-pareto}) for prospective wet-lab validation. The candidates are tested alongside a conventional Benchmark and the leading GEMORNA design GMR-FL-F5~\cite{zhang2025deep} in three independent Fluc2P experiments at 12, 24, 36, 48, and 72\,h after transfection. Consistent with the GEMORNA study, the Benchmark construct comprises the pGL4.11 CDS flanked by the 5' and 3' UTRs of BNT162b2~\cite{polack2020safety}.

\begin{figure}[t]
  \centering
  \includegraphics[width=0.95\linewidth]{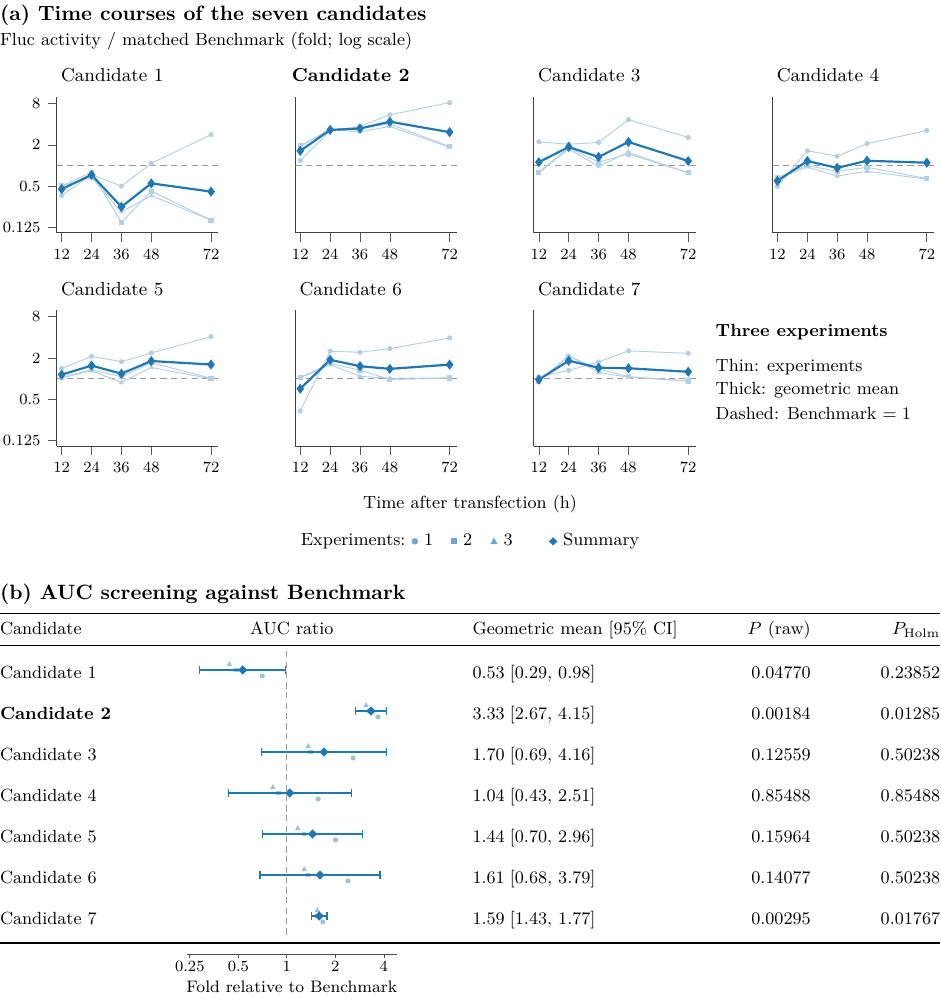}
  \caption{Screening ProMORNA candidates against Benchmark in three independent Fluc2P experiments.}
  \label{fig:wet_screen}
\end{figure}

\Cref{fig:wet_screen}a illustrates the Fluc activity normalized to the benchmark at each time point and \cref{fig:wet_screen}b shows the AUC ratios of each candidate to Benchmark. Candidates~2 and~7 showed higher AUCs after seven-candidate Holm correction: Candidate~2 had a geometric-mean ratio of 3.33 (95\% CI, 2.67--4.15; $P_{\mathrm{Holm}}=0.01285$), and Candidate~7 had a ratio of 1.59 (1.43--1.77; $P_{\mathrm{Holm}}=0.01767$). Candidate~2 is thus selected for closer comparison with the controls.

\begin{figure}[htbp]
  \centering
  \includegraphics[width=\linewidth]{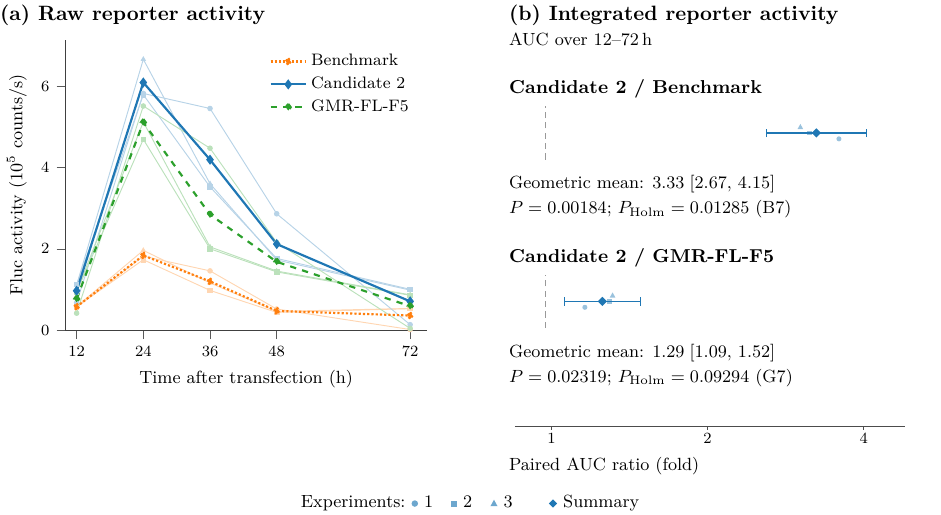}
  \caption{Candidate~2 compared with Benchmark and GMR-FL-F5 in the same three experiments. Holm-adjusted $P$ values are shown for the corresponding seven-candidate families: B7 for Benchmark and G7 for GMR-FL-F5.}
  \label{fig:wet_lead}
\end{figure}

\Cref{fig:wet_lead}a shows the raw activity trajectories of Candidate~2 and the two controls. Candidate~2 is numerically above both controls at every sampled time point in each experiment, peaking at 24\,h and declining by 72\,h. Using the same experiments, its AUC ratio relative to GMR-FL-F5 was 1.29 (95\% CI, 1.09--1.52; unadjusted $P=0.02319$; $P_{\mathrm{Holm}}=0.09294$; \cref{fig:wet_lead}b). This estimated 28.7\% gain remains exploratory after correction across the seven GMR-FL-F5 comparisons.

\FloatBarrier

\section{Conclusion}

In this paper, we propose \textbf{ProMORNA}, an end-to-end generative framework for full-length mRNA design. ProMORNA generates complete mRNA transcripts, including the 5' UTR, CDS, and 3' UTR, directly from a target protein sequence. Using a BART-style encoder--decoder architecture, the model conditions every generated region on the full protein context, enabling joint design of coding and non-coding regions without requiring template inputs or updating model weights at inference time.

Following a progressive training pipeline that combines large-scale evolutionary pre-training with two-stage post-training, ProMORNA aligns its generative distribution with therapeutic mRNA design requirements. Unlike conventional scalar-reward optimization, our RL post-training computes and stabilizes sequence advantages at the level of individual biological metrics prior to aggregation. This formulation mitigates the instability caused by objectives with heterogeneous scales and shifting variances, providing a flexible mechanism for balancing competing design constraints.

\textit{In silico} experiments on Fluc2P demonstrate that ProMORNA improves generated mRNA candidates over supervised baselines and GEMORNA, particularly on the two primary predicted objectives: predicted stability and predicted translation efficiency. Moreover, Pareto frontier analysis confirms that our post-training optimizes the model toward generating mRNA candidates with better multi-objective trade-offs. In the \textit{in vitro} screen, Candidates~2 and~7 show significantly higher time-integrated reporter activity than Benchmark after Holm adjustment. These results demonstrate that ProMORNA is capable of generating functionally active full-length mRNA candidates.

A primary limitation of the current study is that our prospective biological validation was restricted to seven Pareto-selected candidates for a single reporter protein, Fluc2P. The retrospective AUC analysis and prioritization of Candidate~2 require independent validation, particularly against GMR-FL-F5. Reporter activity does not directly measure mRNA stability or translation efficiency. Extensive \textit{in vitro} experiments with longer transcripts across diverse therapeutic contexts are required to rigorously establish the generalizability and practical applicability of ProMORNA. This future work will provide a stronger foundation for applying ProMORNA to therapeutic mRNA design. Moreover, improving the accuracy and experimental robustness of the stability and translation efficiency predictors will be essential for more reliable reward modeling.

\printbibliography[heading=bibintoc]

\clearpage
\begin{appendices}
\crefalias{section}{appendix}
\crefalias{subsection}{appendix}

\section{Experimental Setup}
\label{appendix:exp-setup}

\subsection{Model Training}
\label{appendix:model-training-setup}

\paragraph{Supervised training details.}
The BASE model is trained with the scalable Muon variant proposed in Moonlight~\cite{liu2025muon}, which augments the original Muon optimizer~\cite{jordan2024muon} with weight decay and parameter-wise update-scale adjustment. We use learning rate $10^{-3}$, weight decay $0.1$, equivalent batch size $1{,}024$, and linear warmup over $1{,}000$ steps followed by cosine annealing to $1\%$ of the peak learning rate. We train for $20$ epochs and use the final checkpoint as BASE; SFT is initialized from the BASE checkpoint and trained on 37,369 NM-prefixed RefSeq mRNAs without upstream open reading frames (uORFs), with batch size $128$ and peak learning rate $5\times10^{-5}$ for $3$ epochs.

\paragraph{RL post-training details.}
ProMORNA-RL is initialized from the SFT checkpoint, and the reference policy $\pi_{\mathrm{ref}}$ is fixed as a frozen copy of the SFT model. Then we employ our stabilized MO-GRPO post-training procedure (\cref{sec:mo-grpo}) as a general protein-conditioned optimization stage: we randomly sample $4{,}000$ protein prompts from the training set (firefly luciferase target excluded), pulling a batch of $4$ protein prompts from this pool at each RL step. For each prompt, the policy generates a group of $G=32$ complete mRNA rollouts for policy updates. We train for $1{,}000$ RL steps using AdamW~\cite{loshchilov2019decoupled} with constant learning rate $10^{-5}$ and linear warmup over $50$ steps. The KL penalty coefficient is set to $\beta=0.02$ and the advantage clipping bound is set to $c=3.0$. The policy ratio clipping term is naturally omitted since we only perform one optimizer step per batch. The objective weights are set to $(w_{\mathrm{stability}}, w_{\mathrm{TE}}, w_{\mathrm{MFE}}, w_{\mathrm{U}}, w_{\mathrm{UTR}})=(1.0,1.0,-0.5,-0.5,0.5)$, where negative weights indicate minimization objectives. The damping factors are set to $(\delta_{\mathrm{stability}}, \delta_{\mathrm{TE}}, \delta_{\mathrm{MFE}}, \delta_{\mathrm{U}}, \delta_{\mathrm{UTR}})=(0.2,0.2,0.01,0.01,0.01)$. Invalid sequences receive the maximal penalty before post-centralization. The stability and translation efficiency predictors are frozen throughout RL training.

\paragraph{Reward predictor training.}
Predicted stability and predicted translation efficiency serve as the two primary functional objectives in our reward design. Since these quantities cannot be directly computed from sequence alone, we train two supervised proxy predictors to estimate them from mRNA sequences. Each mRNA sequence is first encoded by a frozen mRNABERT model~\cite{xiong2025mrnabert}, and the resulting token-level representations are aggregated into a 768-dimensional sequence representation by global average pooling. This reduces the two prediction tasks to independent mappings from mRNABERT sequence embeddings to scalar property values. The translation-efficiency and stability predictors are then trained with ridge regression on human measurements from the transcriptome-wide atlas~\cite{zheng2025predicting} and the mammalian mRNA degradation compendium~\cite{agarwal2022genetic}, respectively.

The ridge regularization strength $\alpha$ is optimized separately for each task using Optuna~\cite{akiba2019optuna}. For each predictor, we run 100 hyperparameter trials and sample $\alpha$ from the logarithmic range $[10^{-2}, 10^5]$. The hyperparameter search primarily maximizes validation Spearman's rank correlation coefficient (SRCC). When two trials have validation SRCC values differing by no more than $10^{-4}$, we select the one with the lower validation mean absolute error (MAE). To reduce variance in hyperparameter selection, we use repeated 5-fold cross-validation with three repeats.

\begin{table}[htbp]
  \centering
  \begin{tabular}{lccc}
    \toprule
    Prediction task & SRCC & MAE & $\alpha$ \\
    \midrule
    Stability        & 0.635 & 0.608 & 134.27 \\
    Translation efficiency & 0.744 & 0.364 & 82.80 \\
    \bottomrule
  \end{tabular}
  \caption{Cross-validation performance of the mRNABERT-based proxy predictors.}
  \label{tab:prediction}
\end{table}

\subsection{\textit{In Silico} Experimental Setup}

\paragraph{Target protein.}
For a case study, we evaluate all methods on firefly luciferase (Fluc2P), following the reporter choice in the GEMORNA study~\cite{zhang2025deep}. Only the amino-acid sequence of the target protein is provided to ProMORNA at inference time, and no wild-type mRNA template is used. To avoid potential data leakage, transcripts corresponding to the target protein, if present, are removed from all training datasets and the RL post-training prompt pool. It is worth noting that the framework is protein-agnostic and can generate full-length mRNAs for any given protein sequence.

\paragraph{Baselines.}
We compare ProMORNA-RL with three baselines. BASE denotes the supervised encoder--decoder model trained on the full RefSeq training set. SFT denotes the BASE model further fine-tuned on the filtered \texttt{NM\_} subset. The GEMORNA baseline is constructed by combining the 12 GEMORNA-designed 5' UTRs, the four GEMORNA firefly-luciferase CDSs, and the 10 GEMORNA-designed 3' UTRs from their released supplementary data~\cite{zhang2025deep}. Their Cartesian product yields $12 \times 4 \times 10 = 480$ full-length GEMORNA mRNA candidates. Notably, the GEMORNA-derived sequences used for comparison are selected through wet-lab screening in their original experimental pipeline, and thus represent a carefully selected, high-performing set. For a fair comparison, all candidates from different models are evaluated using the same validity checker and the same evaluation metrics.

\paragraph{Candidate generation.}
For the three ProMORNA variants, sequences are generated using constrained stochastic decoding with temperature $\tau=1.0$, top-$p$ filtering disabled, and a maximum decoder length of $1{,}024$ tokens. Each sequence is forced to begin with \texttt{AGG} to ensure fair comparison with other methods.
We repeat this generation procedure until $1{,}000$ valid candidates are produced. A generated sequence is accepted only if it passes these basic biological filters:
\begin{itemize}
  \item $|\text{5' UTR}| \geq 20$ and $|\text{3' UTR}| \geq 30$;
  \item the CDS correctly translates to the target protein;
  \item no uORF is present, i.e. $\texttt{"AUG"} \notin \text{5' UTR}$.
\end{itemize}

For GEMORNA, $480$ candidates are constructed from the Cartesian product of its released 5' UTR, CDS, and 3' UTR elements.

\paragraph{Evaluation protocol.}
All reported metric distributions are computed on valid and unique candidates. For each method, we evaluate (i) reward-related metrics, including predicted stability, predicted translation efficiency, normalized MFE, U-content, 5' UTR length, and 3' UTR length; (ii) diagnostic sequence-property metrics, including GC content, Codon Adaptation Index (CAI), the normalized MFE of the 5' leader region, the normalized MFE of the remaining mRNA body, external GEMORNA UTR scores, and 5' and 3' UTR Max Identity. For RL learning curves, we report the average metric value over all valid generated candidates in every training step.

\subsection{\textit{In Vitro} Validation}
\label{appendix:in-vitro}

\paragraph{Plasmid Construction.}
The plasmids used in this study are constructed via restriction enzyme cloning and Gibson assembly~\cite{gibson2009enzymatic}.

\paragraph{Reaction Assembly.}
In vitro transcription (IVT) reactions are assembled as described in \cref{tab:ivt_mixture}.
After brief centrifugation, the reaction is incubated at 37$^\circ$C for 1.5--2 hours.

\begin{table}[ht]
  \centering
  \begin{tabular}{lc}
    \toprule
    \textbf{Component} & \textbf{Amount} \\
    \midrule
    RNase-free H$_2$O & Up to 20 $\mu$L \\
    10$\times$ Transcription Buffer & 2 $\mu$L \\
    ATP / CTP / GTP / N1-Me-pUTP (100 mM each) & 2 $\mu$L each \\
    CAP GAG 3011 (100 mM) & 2 $\mu$L \\
    Template DNA & 1 $\mu$g \\
    T7 RNA Polymerase Mix & 2 $\mu$L \\
    \midrule
    \textbf{Total Reaction Volume} & \textbf{20 $\mu$L} \\
    \bottomrule
  \end{tabular}
  \caption{Composition of the \textit{in vitro} transcription reaction.}
  \label{tab:ivt_mixture}
\end{table}

\paragraph{DNA Digestion.}
DNase I (10 U) is added to each reaction, followed by incubation at 37$^\circ$C for 30 minutes.

\paragraph{RNA Purification.}
The RNA product is purified via lithium chloride precipitation. An equal volume of lithium chloride precipitation solution (5 M) is added to the reaction product and mixed thoroughly. The mixture is precipitated at $-20^\circ$C for 30 minutes and then centrifuged at 12,000 RPM at 4$^\circ$C for 15 minutes. The supernatant is discarded. The RNA pellet is washed with 200 $\mu$L of 70\% ice-cold ethanol and centrifuged at 12,000 RPM at 4$^\circ$C for 15 minutes. The supernatant is discarded and the washing step is repeated 2--3 times. The pellet is air-dried with the lid open for 5--10 minutes. Once completely dry, the RNA pellet is dissolved in 100--200 $\mu$L of RNase-free ddH$_2$O or an appropriate buffer. The RNA concentration is measured, aliquoted, and stored at $-80^\circ$C.

\paragraph{mRNA Transfection.}
HEK293T cells are cultured in DMEM (Gibco, 6126249) supplemented with 10\% FBS (Vazyme, 7E951A5), 100 units/mL penicillin, and 100 $\mu$g/mL streptomycin. One day prior to transfection, cells are seeded into 96-well plates at a density of 20,000 cells per well. Transfection is performed using Lipofectamine\textsuperscript{\textregistered} RNAiMAX Reagent (Invitrogen, 13778-030) following the manufacturer's protocol.

\paragraph{Cell Lysis.}
To measure firefly luciferase activity, 100 $\mu$L of reporter cell lysis buffer is added to each well to ensure complete cell lysis. After lysis, the lysate is centrifuged at 10,000--15,000$\times$g for 3--5 minutes, and the supernatant is collected for subsequent measurement.

\paragraph{Detection Reagent Preparation.}
Firefly luciferase detection reagent is thawed and equilibrated to room temperature. The chemiluminescence reader (or a multifunctional microplate reader with chemiluminescence detection capability) is operated according to the manufacturer's instructions, with appropriate interval and measurement times set based on experimental requirements.

\paragraph{Luminescence Measurement.}
For each sample, 80 $\mu$L of supernatant is transferred to the assay plate. An equal volume of reporter cell lysis buffer serves as a blank control. Subsequently, 100 $\mu$L of firefly luciferase detection reagent is added to each well, mixed thoroughly by pipetting, and the luminescence signal is immediately measured.

\paragraph{Data Processing and Statistical Analysis.}
We average first-cycle readings from technical wells within each construct, experiment, and time point (four wells in experiment 1 and three in experiments 2 and 3). In this retrospective analysis, the principal activity measure is the trapezoidal AUC of the raw counts/s curve over 12--72\,h, calculated before normalization. All five time points are retained, with no additional background subtraction. Candidate/control AUC ratios are paired within experiment. We report geometric means and pointwise 95\% confidence intervals obtained from the three log ratios using the $t$ distribution ($n = 3, df = 2$). These intervals assume approximately normal log ratios and are not adjusted for multiplicity or candidate selection. Two-sided $t$ tests assess whether the mean log ratio differs from zero. Holm correction~\cite{holm1979simple} controls family-wise error separately within the seven candidate--Benchmark comparisons (B7; screening) and the seven candidate--GMR-FL-F5 comparisons (G7; exploratory). Time courses are descriptive.

\section{Reinforcement Learning Dynamics}
\label{appendix:rl-dynamics}

\begin{figure}[htbp]
  \centering

  \begin{subfigure}[b]{0.38\linewidth}
    \centering
    \hspace*{\fill}
    \includegraphics[width=\linewidth]{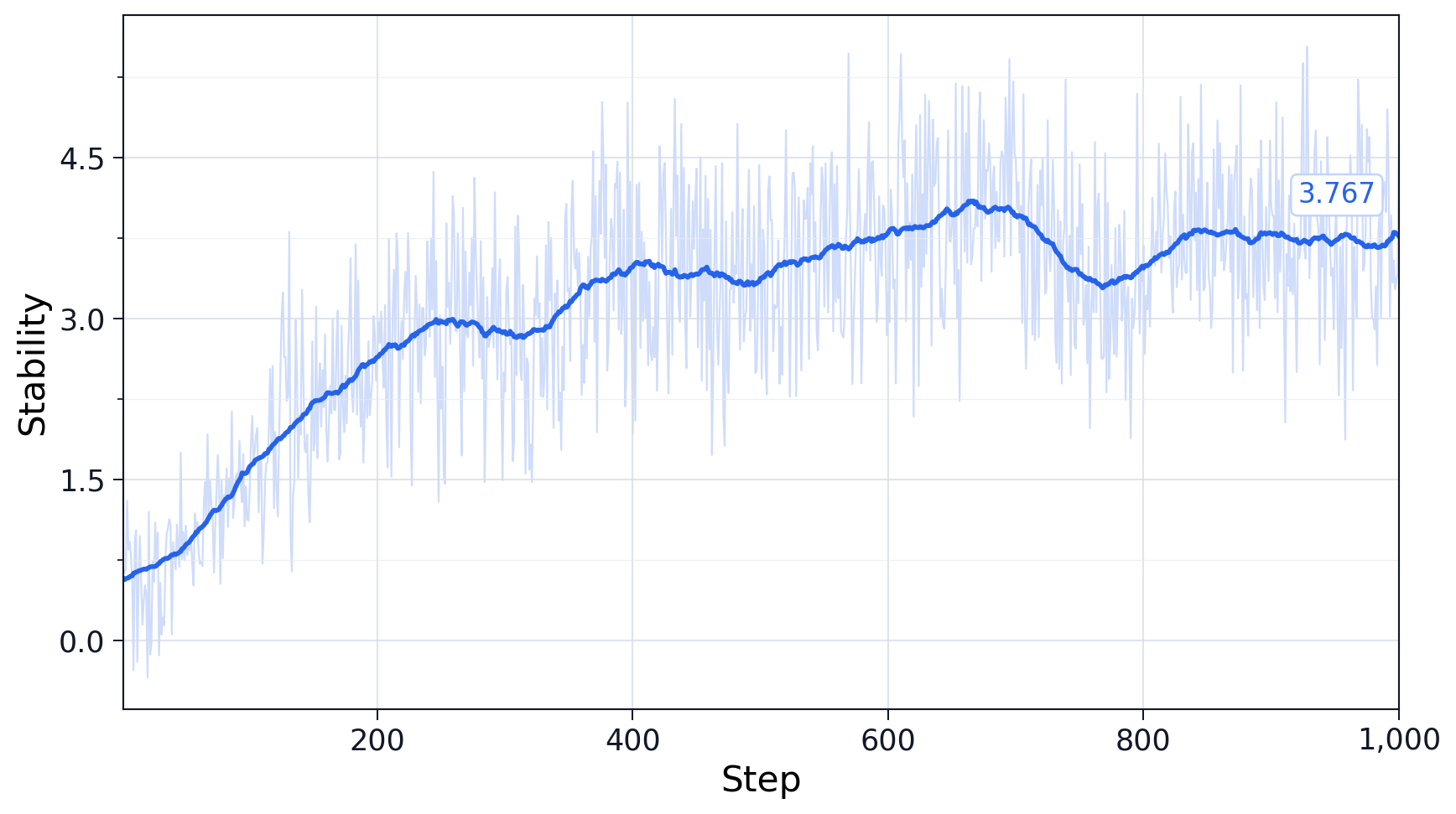}
    \caption{Predicted stability}
  \end{subfigure}\hspace{0.05\linewidth}
  \begin{subfigure}[b]{0.38\linewidth}
    \centering
    \includegraphics[width=\linewidth]{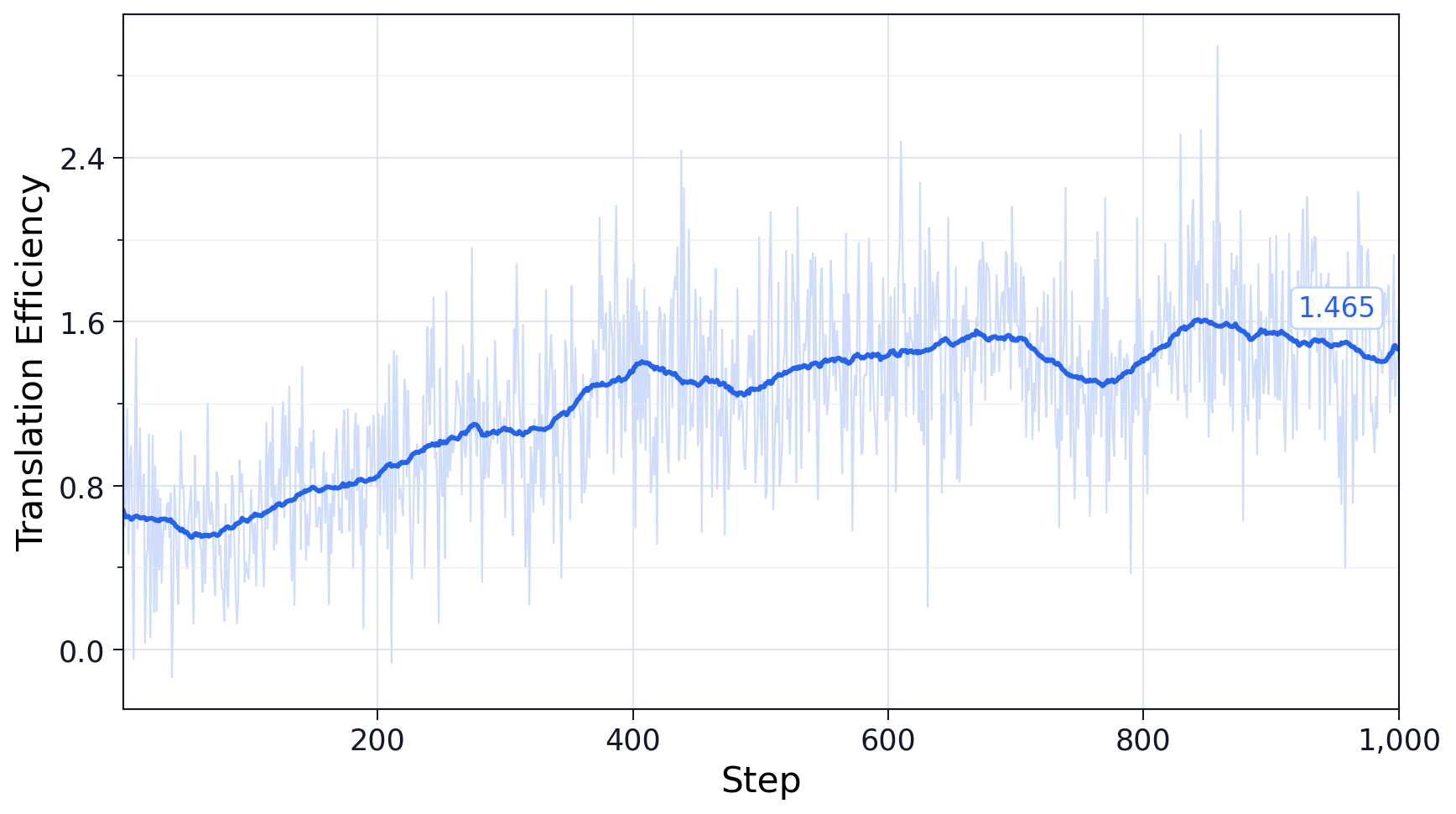}
    \caption{Predicted translation efficiency}
  \end{subfigure}

  \vspace{0.8em}

  \begin{subfigure}[b]{0.3\linewidth}
    \centering
    \includegraphics[width=\linewidth]{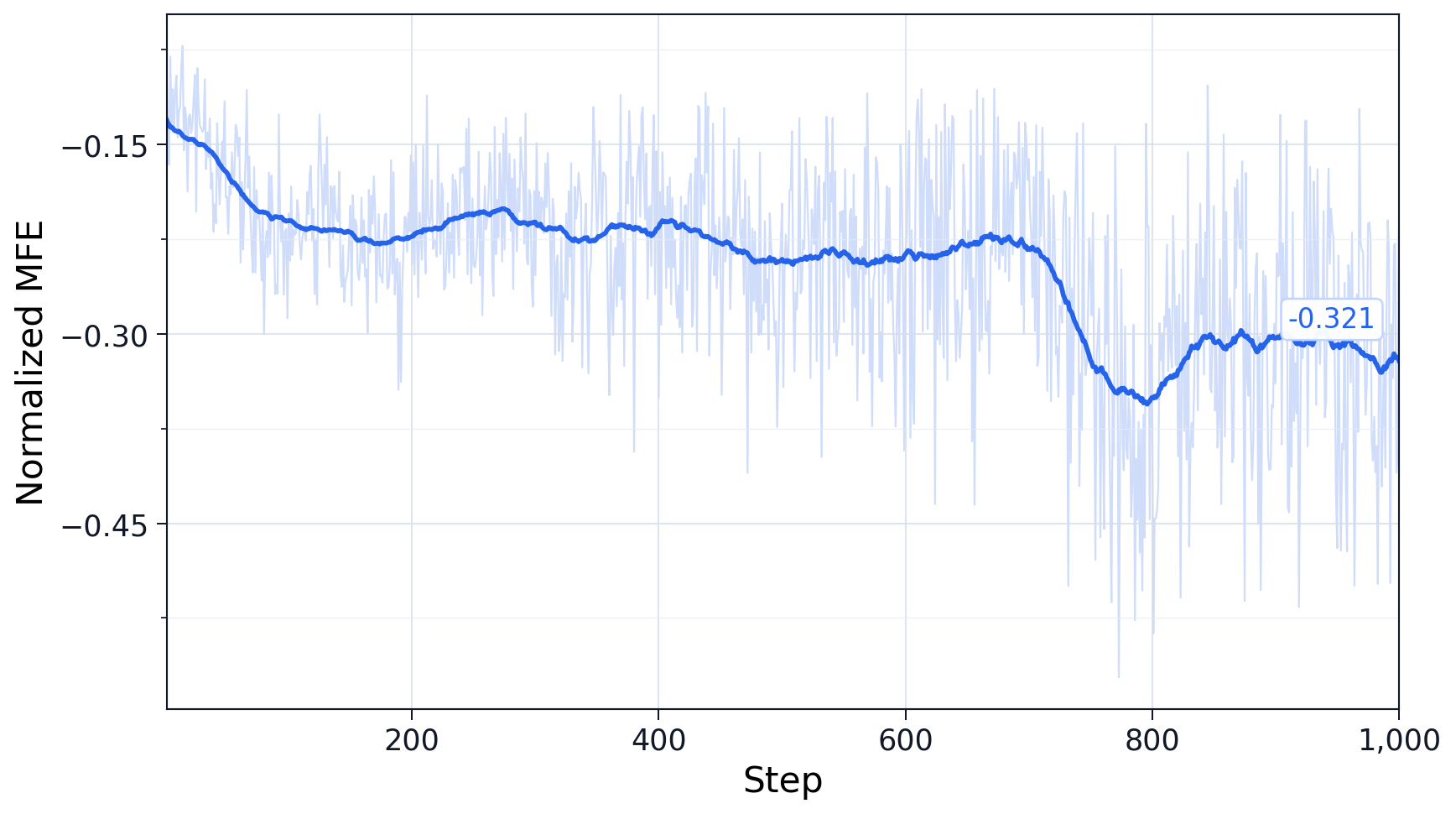}
    \caption{Normalized MFE}
  \end{subfigure}
  \hfill
  \begin{subfigure}[b]{0.3\linewidth}
    \centering
    \includegraphics[width=\linewidth]{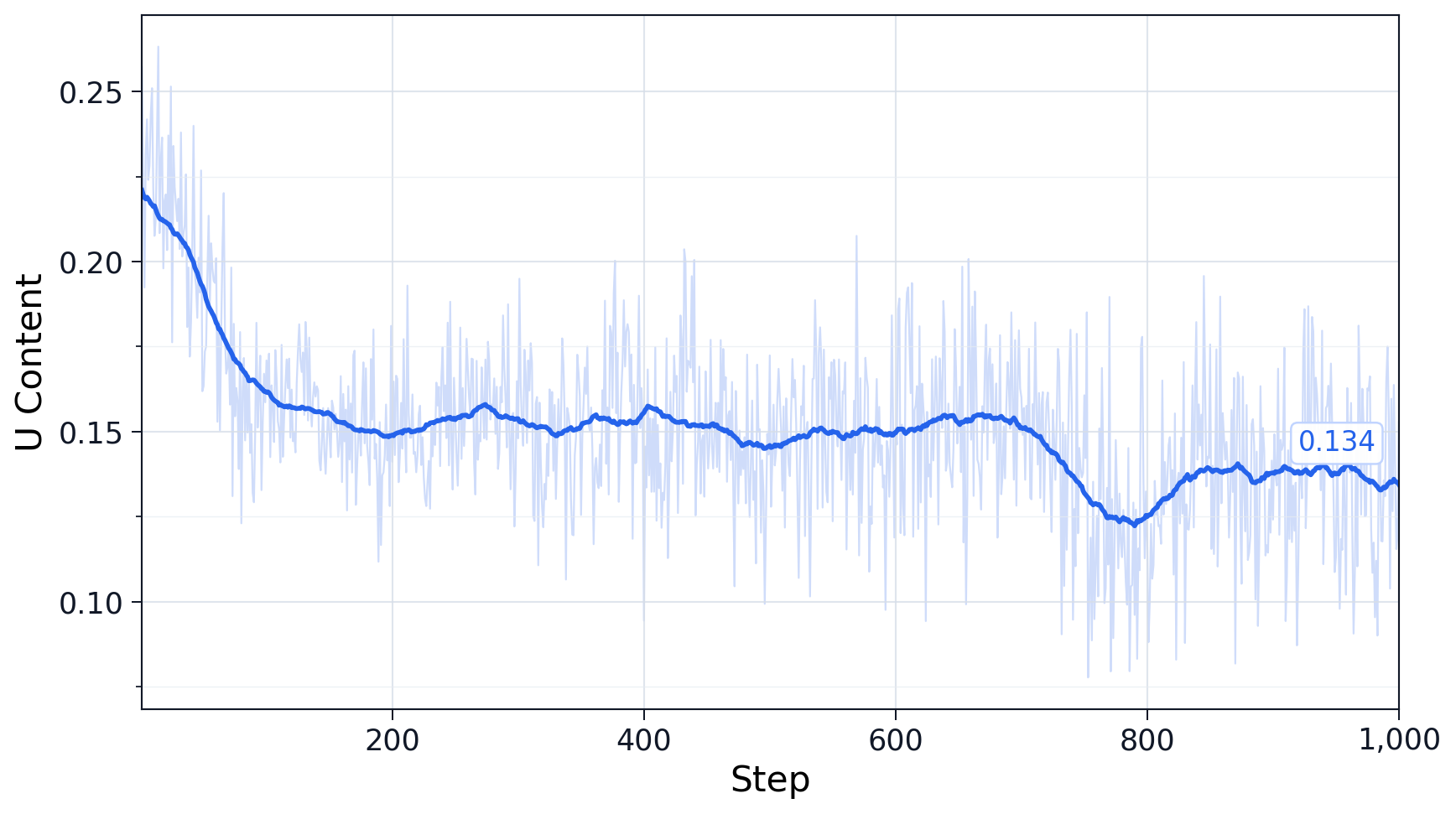}
    \caption{U-content}
  \end{subfigure}
  \hfill
  \begin{subfigure}[b]{0.3\linewidth}
    \centering
    \includegraphics[width=\linewidth]{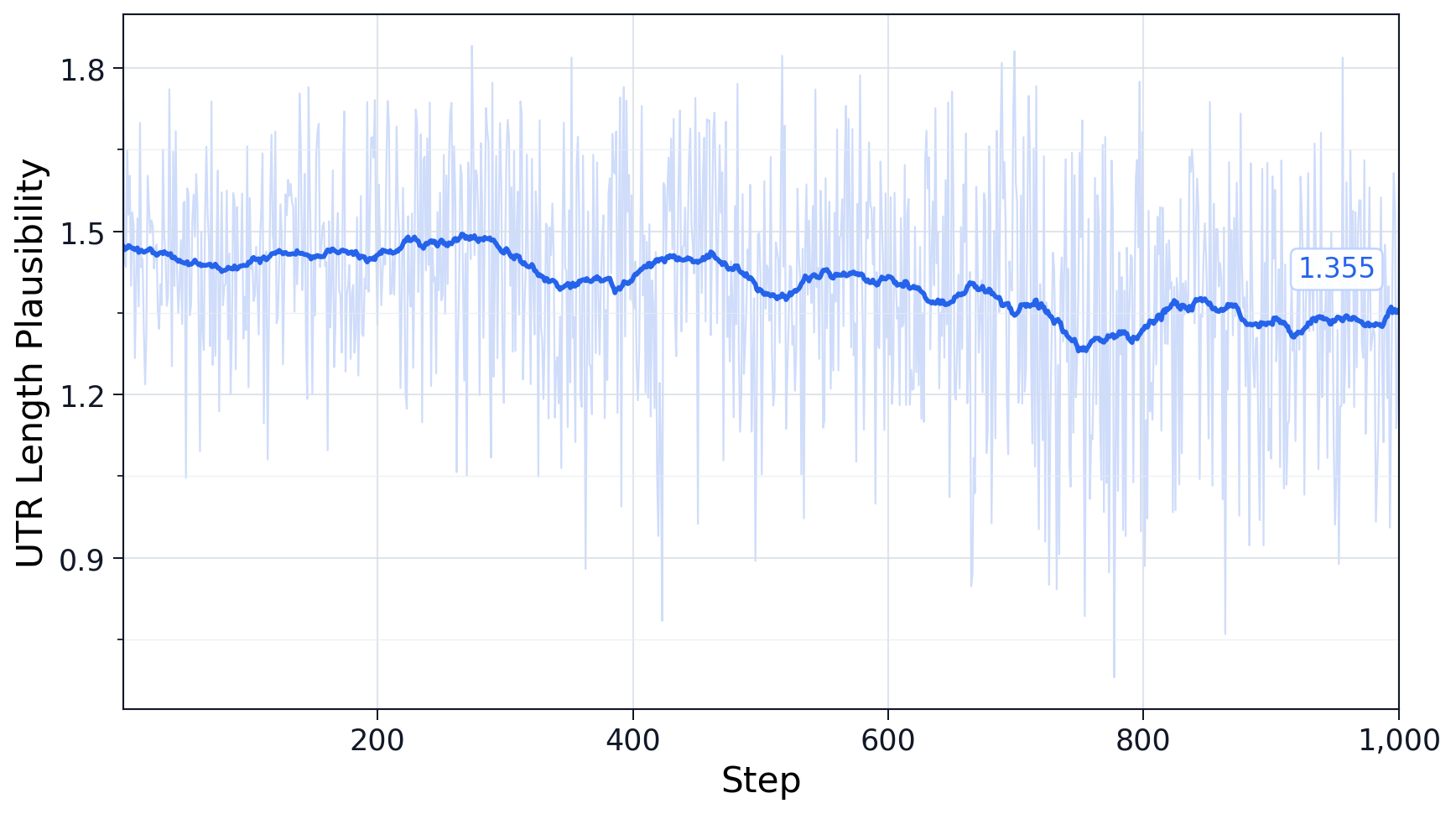}
    \caption{UTR length plausibility}
  \end{subfigure}

  \caption{The learning curves of RL post-training on five reward metrics.}
  \label{fig:learning-curves}
\end{figure}

In this section, we investigate how RL post-training alters the generator. \Cref{fig:learning-curves} presents the learning curves for the five reward-related metrics over 1,000 RL steps, averaged across valid rollouts from protein prompts sampled during general RL post-training. The light traces represent step-level variance, while the bold curves depict the smoothed trends.

RL post-training successfully optimizes the two primary functional proxy objectives. Notably, predicted stability increases rapidly and steadily throughout most of the training. Predicted translation efficiency experiences an initial decline due to early trade-offs but subsequently recovers, rising steadily before gradually approaching a plateau. Concurrently, normalized MFE decreases, indicating the formation of more stable full-transcript secondary structures, and U-content drops sharply, aligning consistently with the reward design. The UTR length plausibility score slightly lowered, which matches our expectations for it as a regularization term. Overall, these trajectories demonstrate that multi-objective RL post-training can effectively improve multiple rewards simultaneously, despite the noise introduced by stochastic full-length rollouts.

\begin{figure}[htbp]
  \centering
  \includegraphics[width=\linewidth]{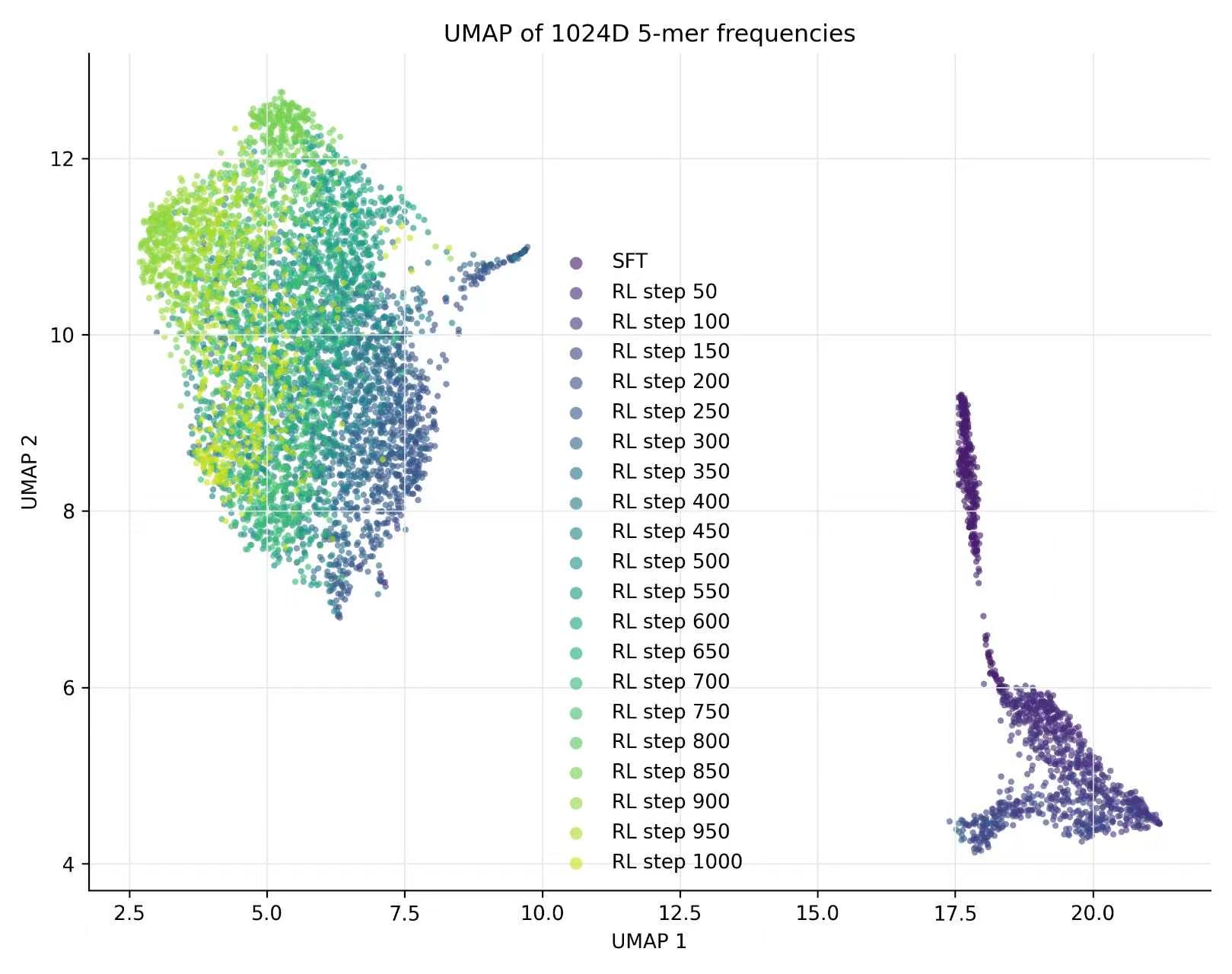}
  \caption{UMAP visualization of 5-mer features from mRNAs generated during RL post-training.}
  \label{fig:umap}
\end{figure}

To understand the evolution of sequence composition during training, \cref{fig:umap} visualizes the generated mRNAs using a UMAP projection~\cite{mcinnes2018umap} of $1{,}024$-dimensional 5-mer frequency features. The projection reveals a striking bifurcation in the latent space: early RL samples cluster tightly around the original SFT baseline on the right, whereas progressing training steps drive the candidates to migrate into a completely separate, vastly expanded region on the left. This distinct spatial shift underscores that RL post-training does not simply exploit the supervised model's local vicinity, but actively and extensively explores novel sequence-composition paradigms. While this compositional shift does not by itself establish biological quality, pairing this structural migration with the previously discussed learning curves demonstrates that on-policy optimization effectively navigates away from the SFT distribution to uncover highly distinct sequences with superior reward profiles.

\begin{figure}[htbp]
  \centering
  \includegraphics[width=\linewidth]{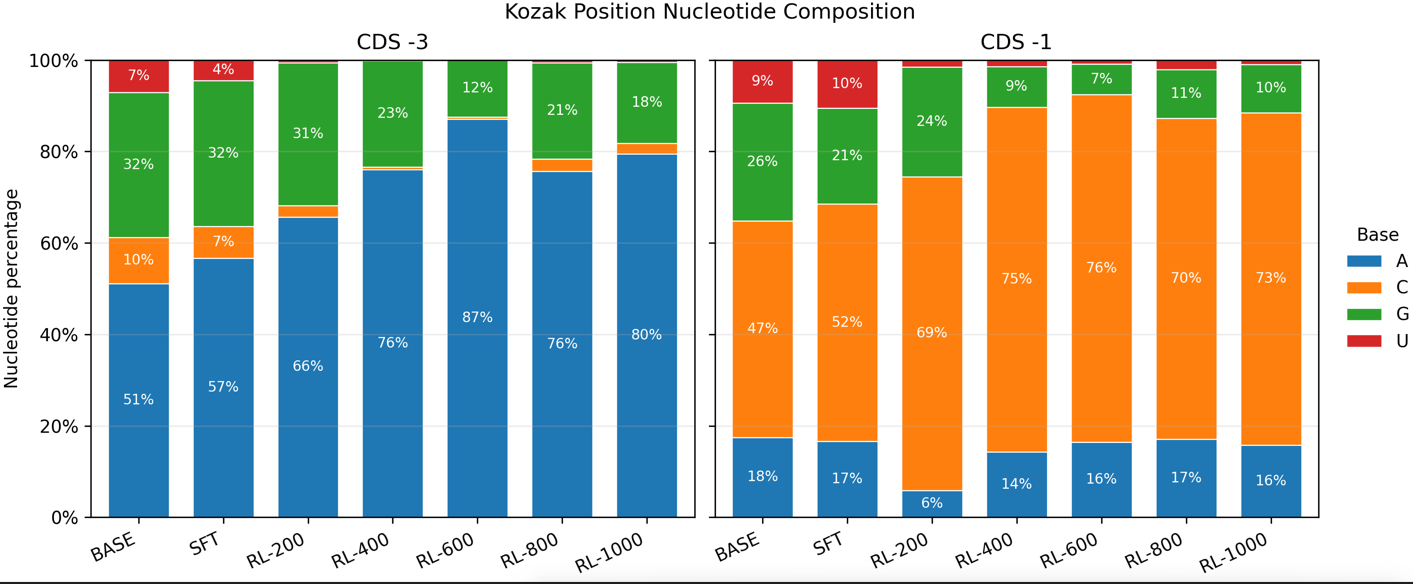}
  \caption{RL shifts nucleotide compositions at Kozak positions.}
  \label{fig:kozak}
\end{figure}

The Kozak sequence is a well-established cis-regulatory element in eukaryotic mRNA that strongly influences translation initiation efficiency~\cite{kozak1987analysis,kozak1986point}. According to the canonical Kozak consensus sequence, the nucleotide at position $-3$ relative to the start codon should preferably be a purine, with adenine (A) being the most optimal, while the $-1$ position is ideally a cytosine (C). We investigated whether our RL-optimized models autonomously learned to respect these biological constraints, despite receiving no explicit Kozak-related inductive bias during training. The results are illustrated in \cref{fig:kozak}.

At the -3 position, the model exhibits a clear progressive enrichment for purines (A/G) across training stages, with a striking shift toward the optimal adenine. The BASE model demonstrates a moderate baseline preference (A+G = 83\%), which is further reinforced by supervised fine-tuning (SFT: 89\%). Strikingly, RL training not only drives the combined purine content near 97--99\% (from RL-400 onward), but it also autonomously discovers the superiority of adenine: the frequency of \texttt{A} alone surges from 57\% in SFT to a peak of 87\% at RL-600, while \texttt{G} concurrently drops from 32\% to 12\%.

Similarly, at the -1 position, the trend toward Kozak compliance is highly pronounced. The BASE model places a cytosine at this position only 47\% of the time, and SFT improves this to 52\%. However, RL training yields a dramatic shift: C content jumps to 69\% at RL-200, peaks at 76\% at RL-600, and remains robustly high across all subsequent checkpoints. Together, these results provide evidence of biological interpretability within our RL pipeline.

\section{MO-GRPO Ablation}
\label{appendix:mo-grpo-ablation}

\begin{figure}[htbp]
  \centering

  \begin{subfigure}[b]{0.33\linewidth}
    \centering
    \includegraphics[width=0.91\linewidth]{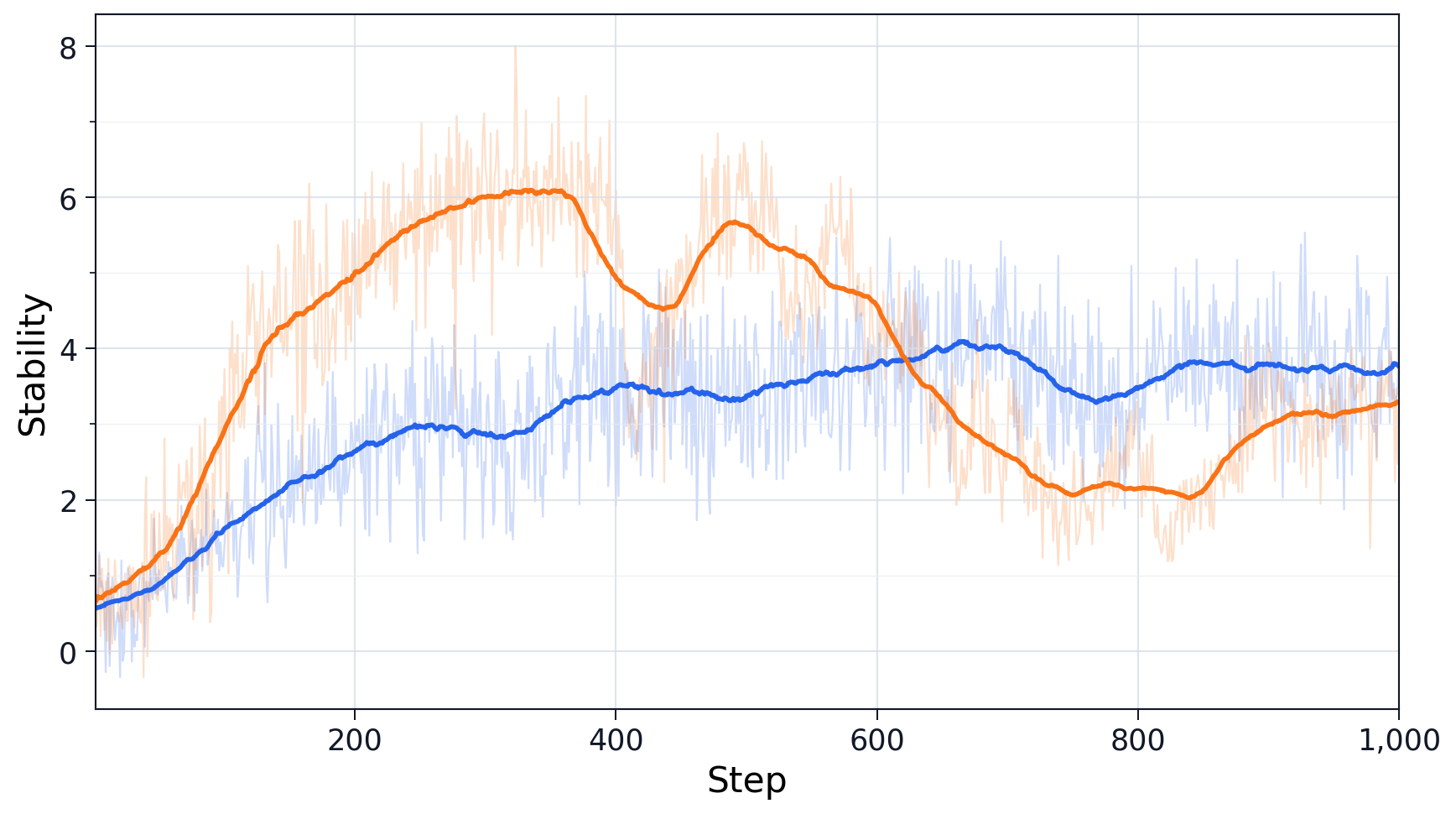}
    \caption{Predicted stability}
    \label{fig:ablation-stability}
  \end{subfigure}%
  \begin{subfigure}[b]{0.33\linewidth}
    \centering
    \includegraphics[width=0.91\linewidth]{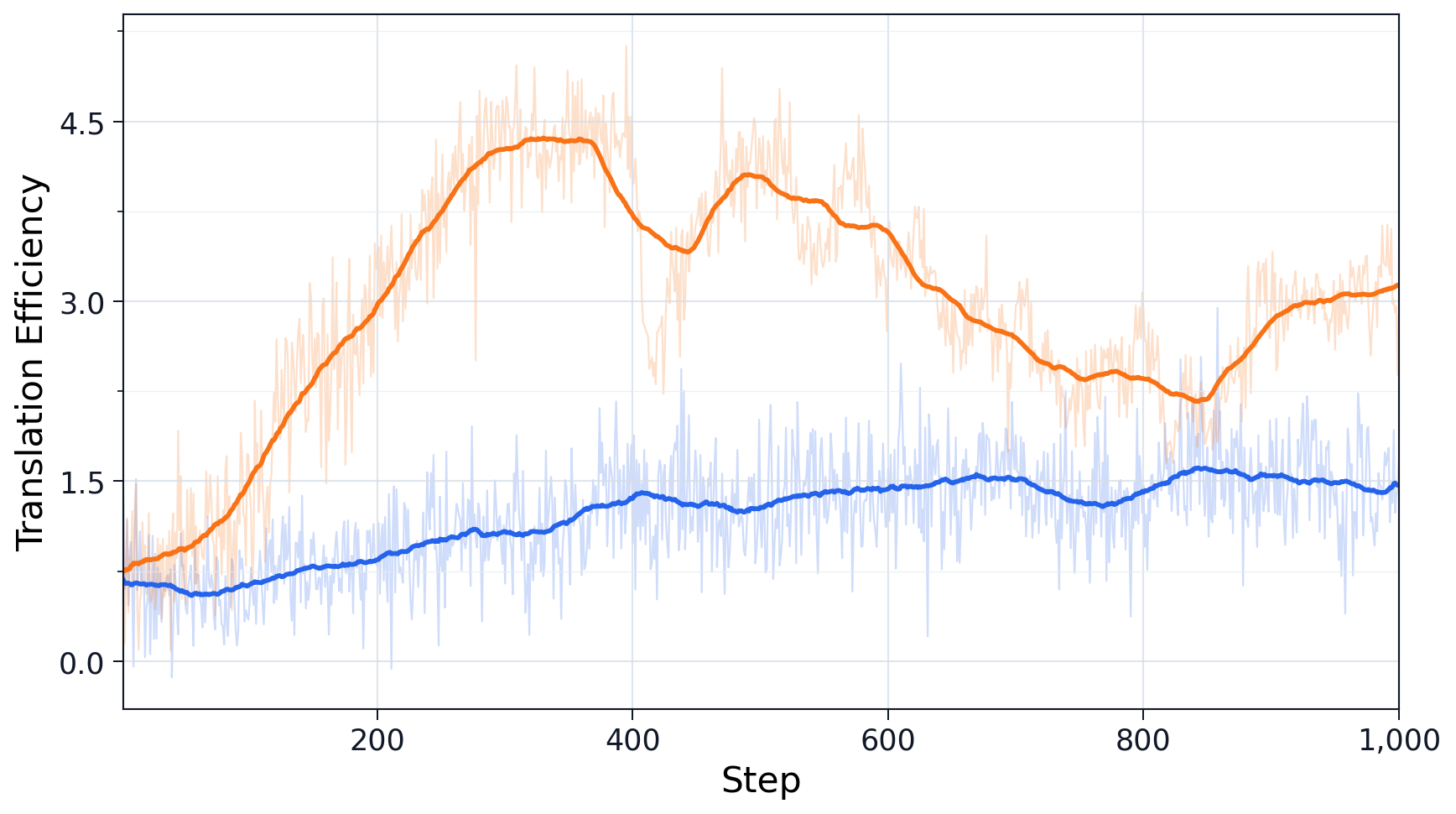}
    \caption{Predicted translation efficiency}
    \label{fig:ablation-te}
  \end{subfigure}%
  \begin{subfigure}[b]{0.33\linewidth}
    \centering
    \includegraphics[width=0.91\linewidth]{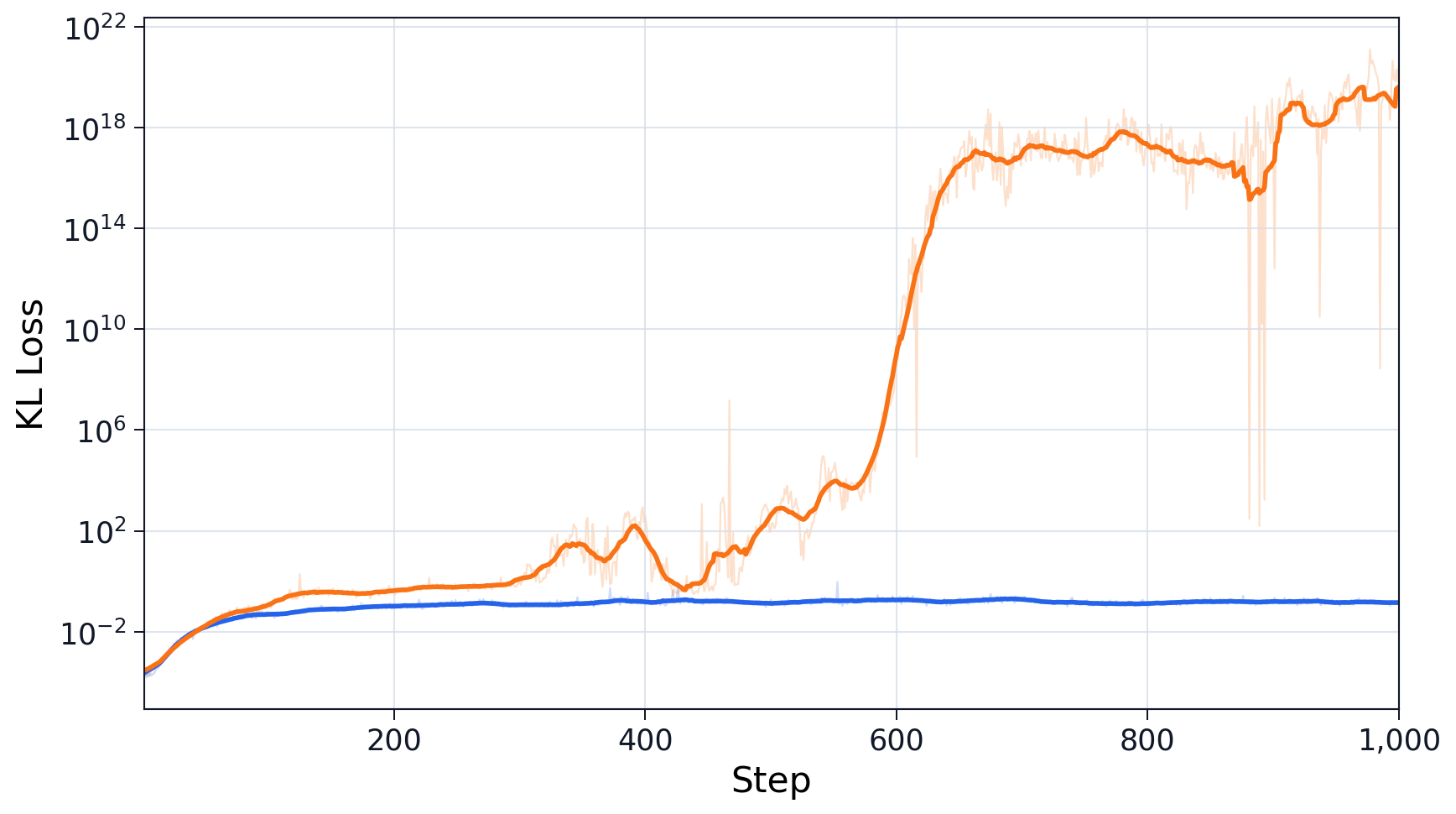}
    \caption{KL penalty loss}
    \label{fig:ablation-kl}
  \end{subfigure}

  \vspace{0.8em}

  \begin{subfigure}[b]{0.33\linewidth}
    \centering
    \includegraphics[width=0.91\linewidth]{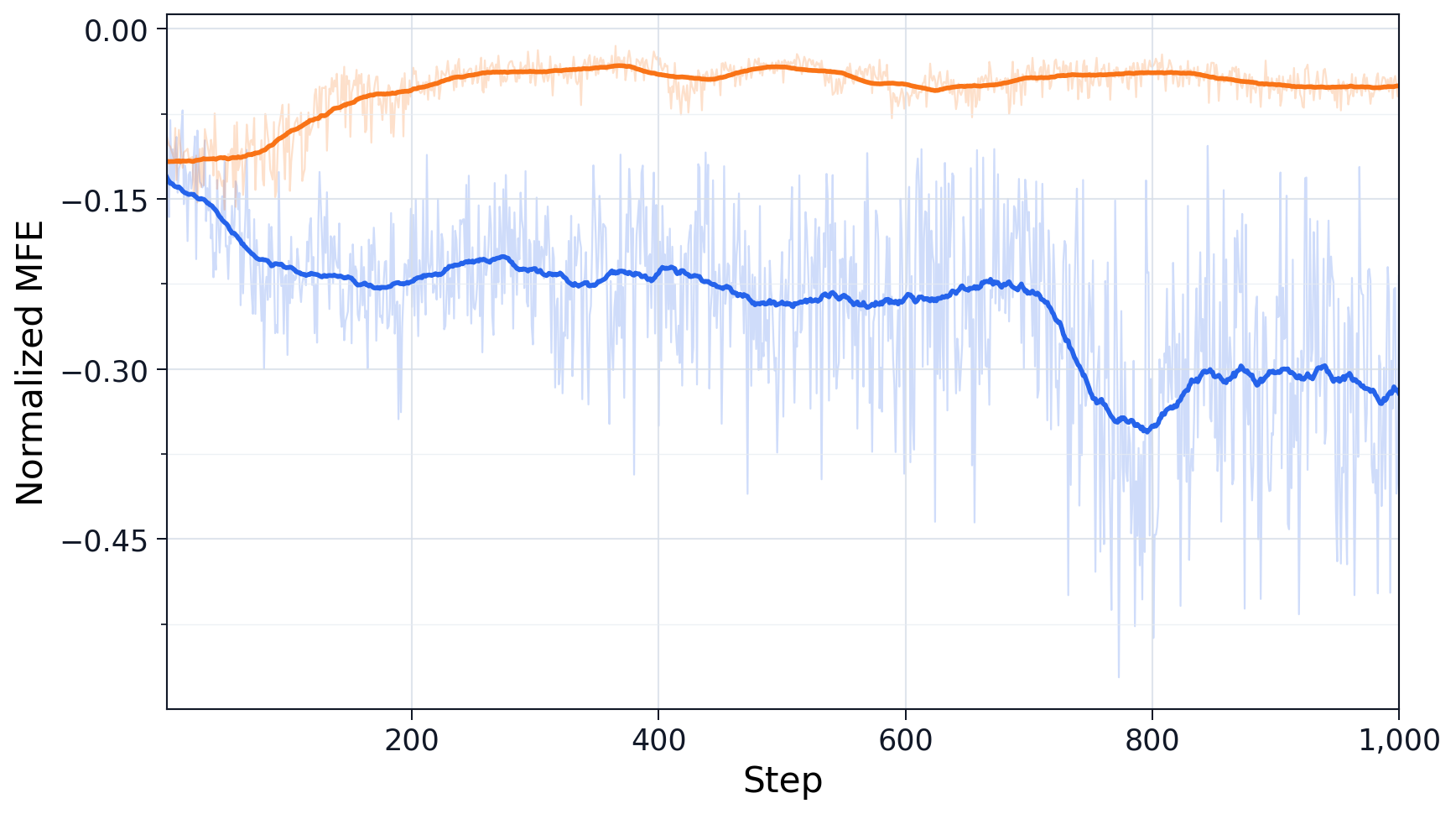}
    \caption{Normalized MFE}
    \label{fig:ablation-mfe}
  \end{subfigure}%
  \begin{subfigure}[b]{0.33\linewidth}
    \centering
    \includegraphics[width=0.91\linewidth]{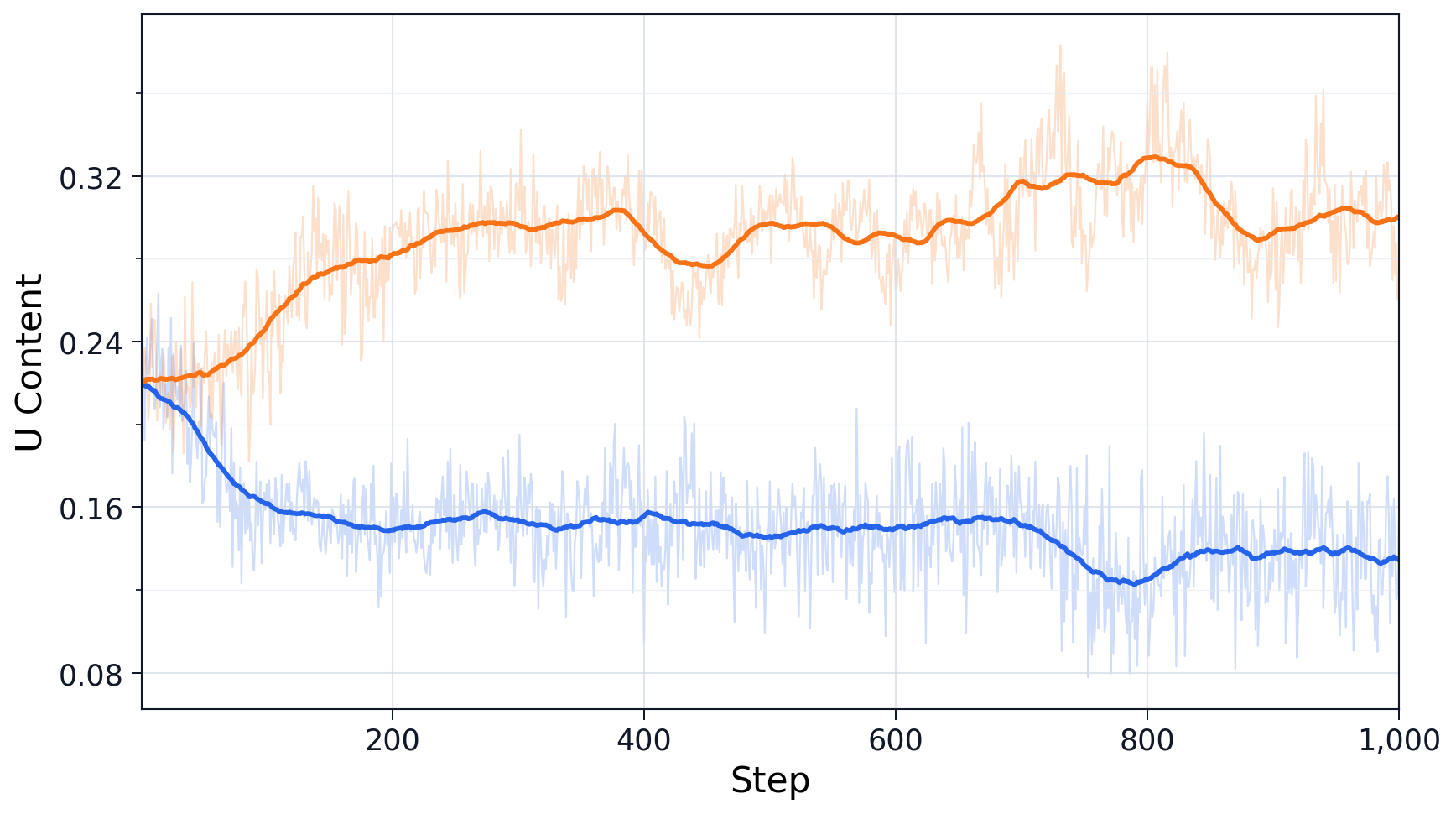}
    \caption{U-content}
    \label{fig:ablation-u-content}
  \end{subfigure}%
  \begin{subfigure}[b]{0.33\linewidth}
    \centering
    \includegraphics[width=0.91\linewidth]{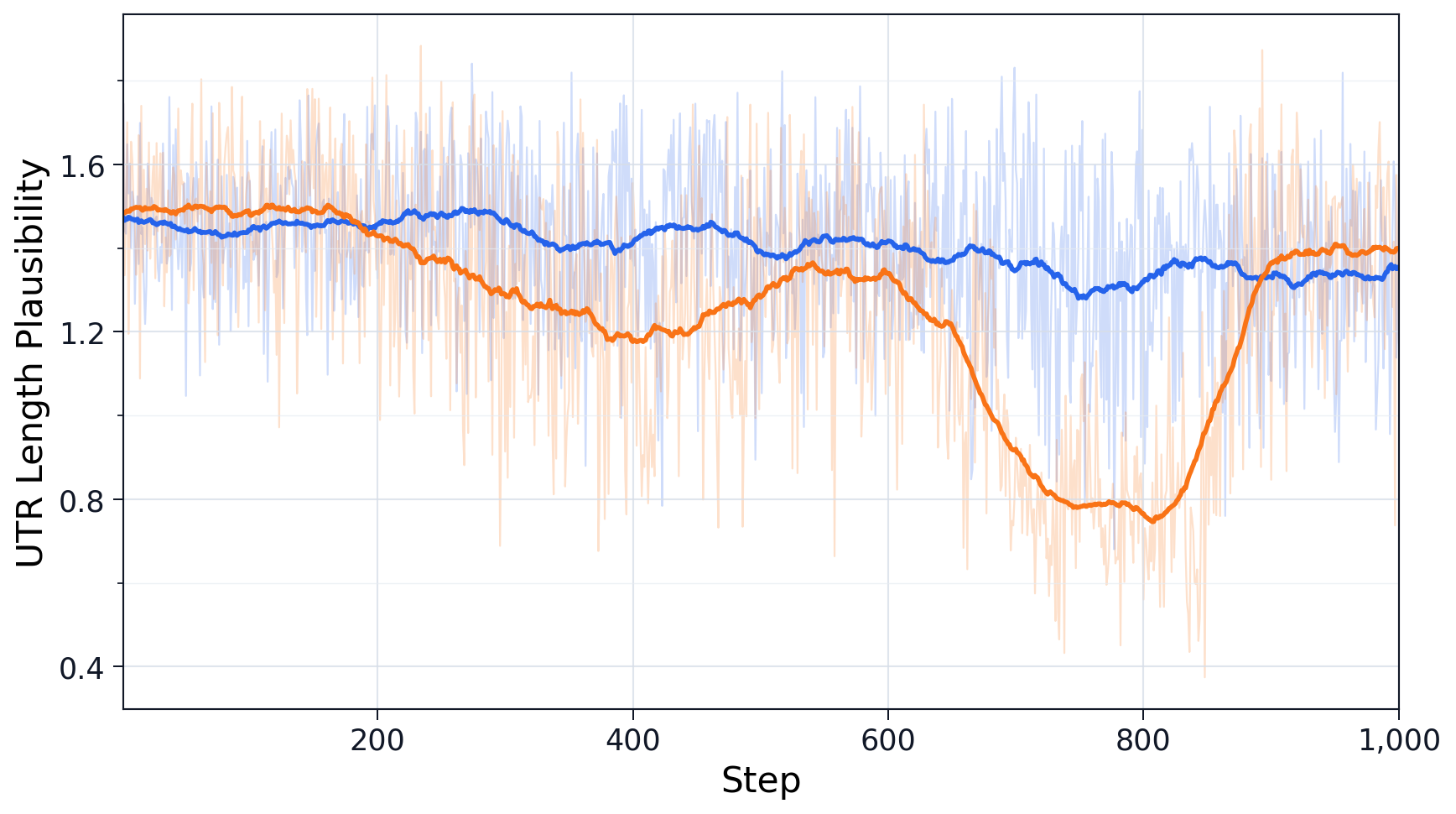}
    \caption{UTR length plausibility}
    \label{fig:ablation-utr-length}
  \end{subfigure}

  \caption{MO-GRPO versus scalarized GRPO on ProMORNA post-training. The orange curves are from scalarized GRPO, while the blue curves are the RL post-training curves reported in \cref{appendix:rl-dynamics}.}
  \label{fig:ablation-curves}
\end{figure}

To evaluate the necessity of metric-level advantage standardization and our stabilization mechanisms, we replace our MO-GRPO formulation in RL post-training with a scalarized GRPO-style ablation, in which the reward is calculated by the weighted sum of the five metrics defined in \cref{sec:metrics}, using exactly the same metric weights reported in \cref{appendix:model-training-setup}, and the advantage is given by standardized scalar rewards. Invalid sequences are assigned a reward of zero. Other hyperparameters remain unchanged. The comparison of the two learning curves is shown in \cref{fig:ablation-curves}.

Consistent with recent theoretical analyses of variance dominance and reward signal collapse in multi-objective GRPO~\cite{ichihara2026mogrpo,liu2026gdpo}, because metric weights aren't divided by their group standard deviations in the scalarized GRPO, predicted stability and translation efficiency dominated the reward calculation at early stages of RL, resulting in severe reward hacking (\cref{fig:ablation-stability,fig:ablation-te}). This caused the policy to diverge largely from the reference SFT policy, and finally training collapsed because of the exploding KL divergence penalty (\cref{fig:ablation-kl}). Due to reward hacking, only the two mRNABERT predictions improved, while all other metrics deteriorated (\cref{fig:ablation-mfe,fig:ablation-u-content,fig:ablation-utr-length}). On the other hand, MO-GRPO training was very stable, benefiting from metric-level advantage calculation~\cite{ichihara2026mogrpo,liu2026gdpo} combined with our dynamic damping, extreme-value clipping, and validity-masked penalty methods.

\section{Additional mRNA Property Analysis}
\label{appendix:additional-analysis}

\begin{figure}[ht]
  \centering

  \begin{subfigure}[b]{0.3\linewidth}
    \centering
    \includegraphics[width=\linewidth]{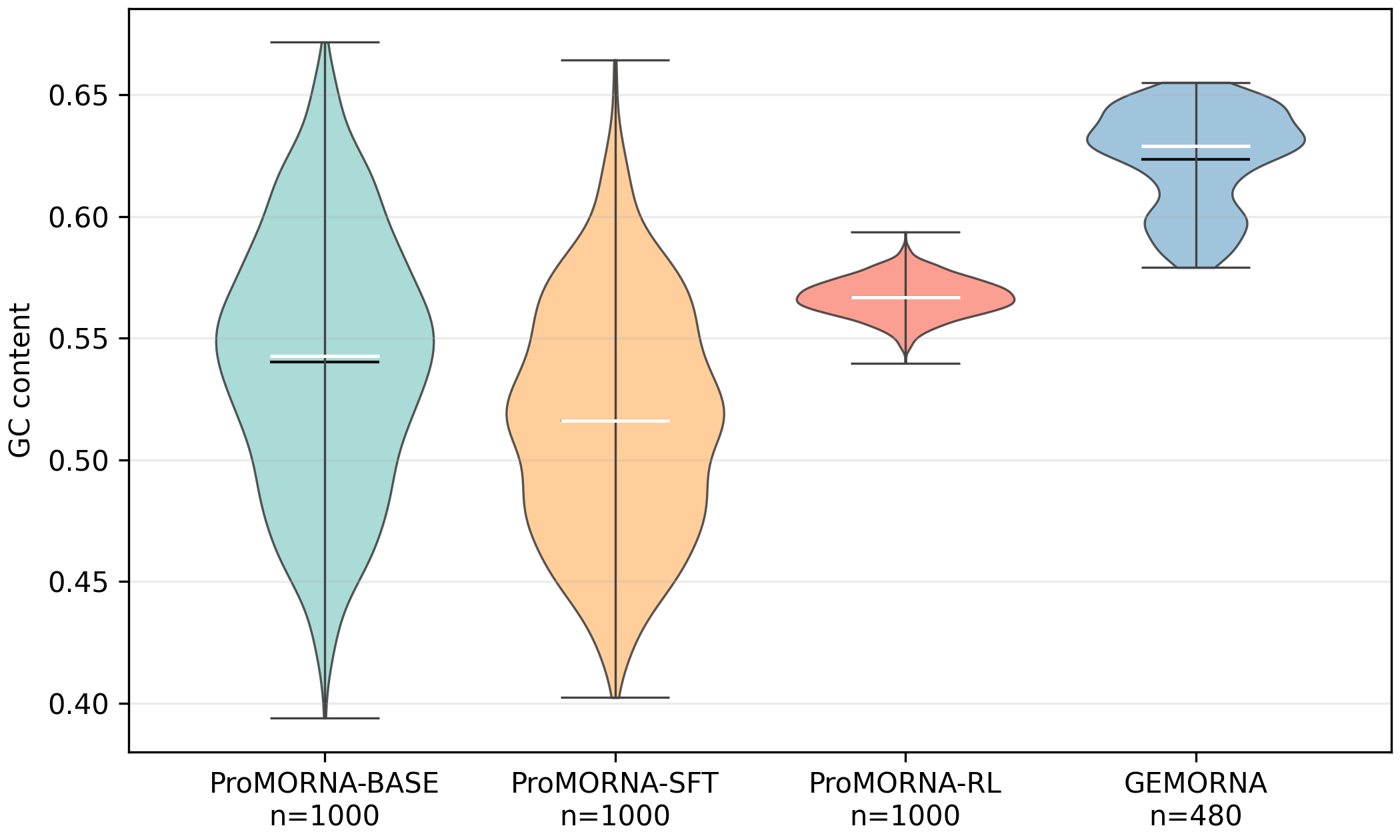}
    \caption{GC content}
    \label{fig:seq_gc}
  \end{subfigure}\hfill
  \begin{subfigure}[b]{0.3\linewidth}
    \centering
    \includegraphics[width=\linewidth]{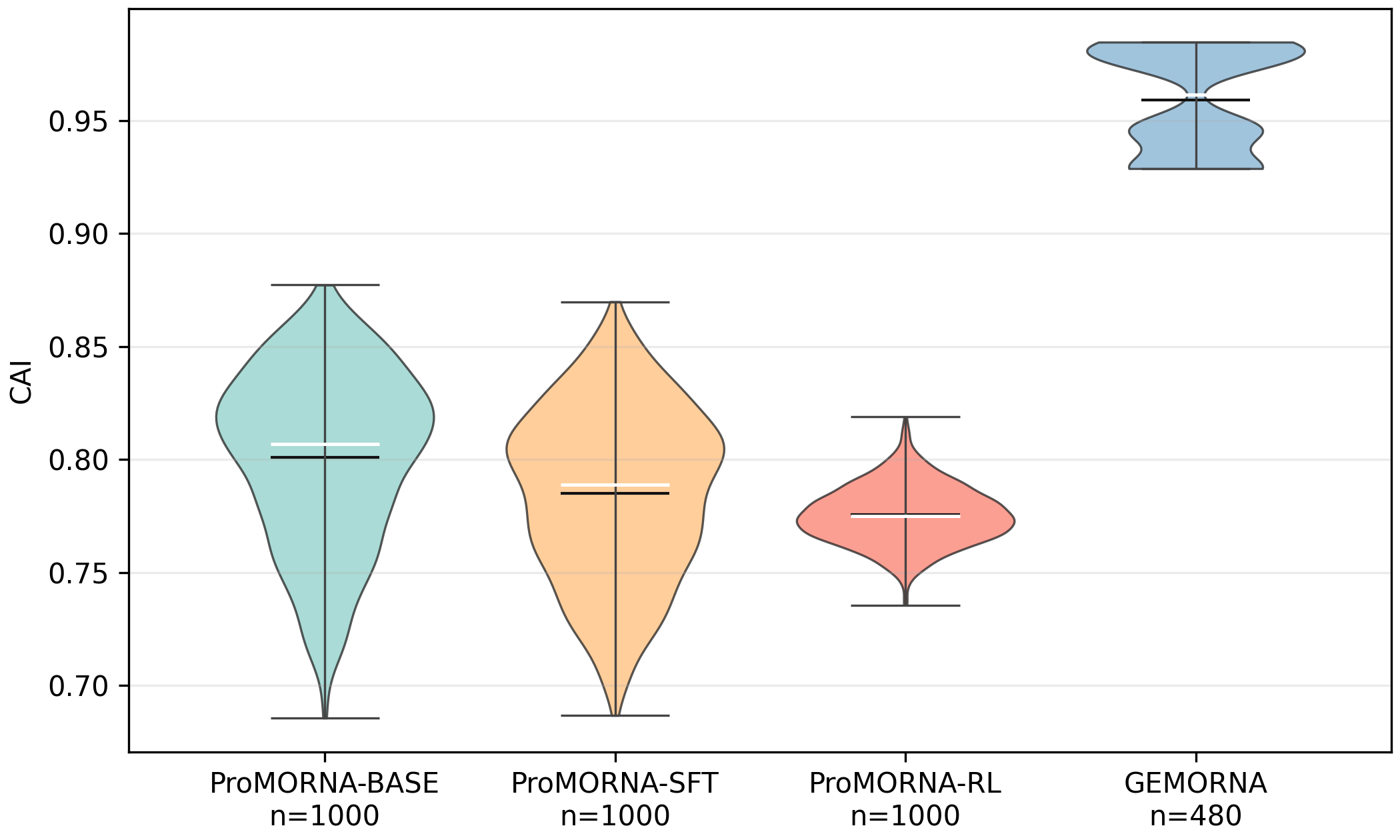}
    \caption{CAI}
    \label{fig:seq_cai}
  \end{subfigure}\hfill
  \begin{subfigure}[b]{0.3\linewidth}
    \centering
    \includegraphics[width=\linewidth]{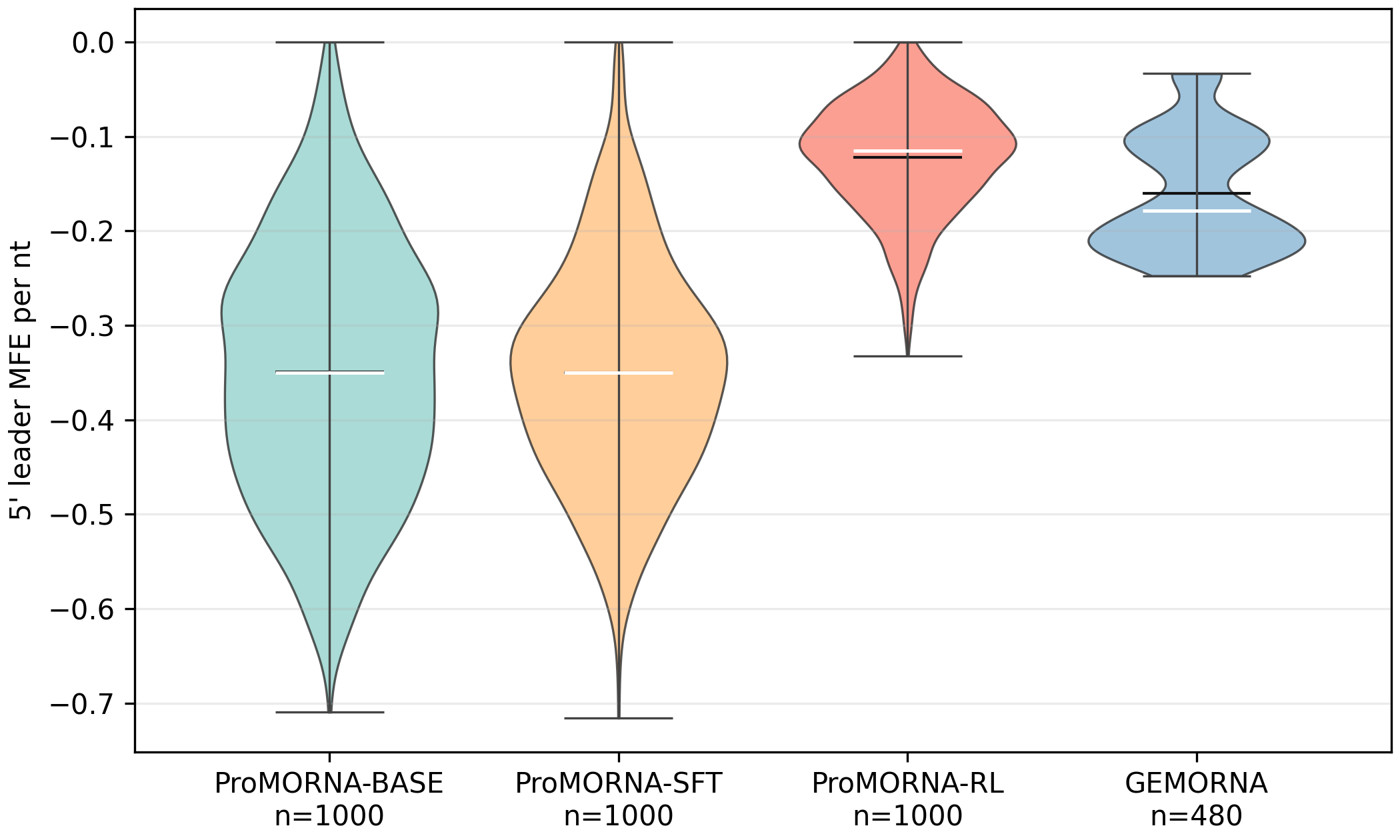}
    \caption{5' leader MFE}
    \label{fig:seq_lmfe}
  \end{subfigure}

  \vspace{0.8em}

  \begin{subfigure}[b]{0.3\linewidth}
    \centering
    \includegraphics[width=\linewidth]{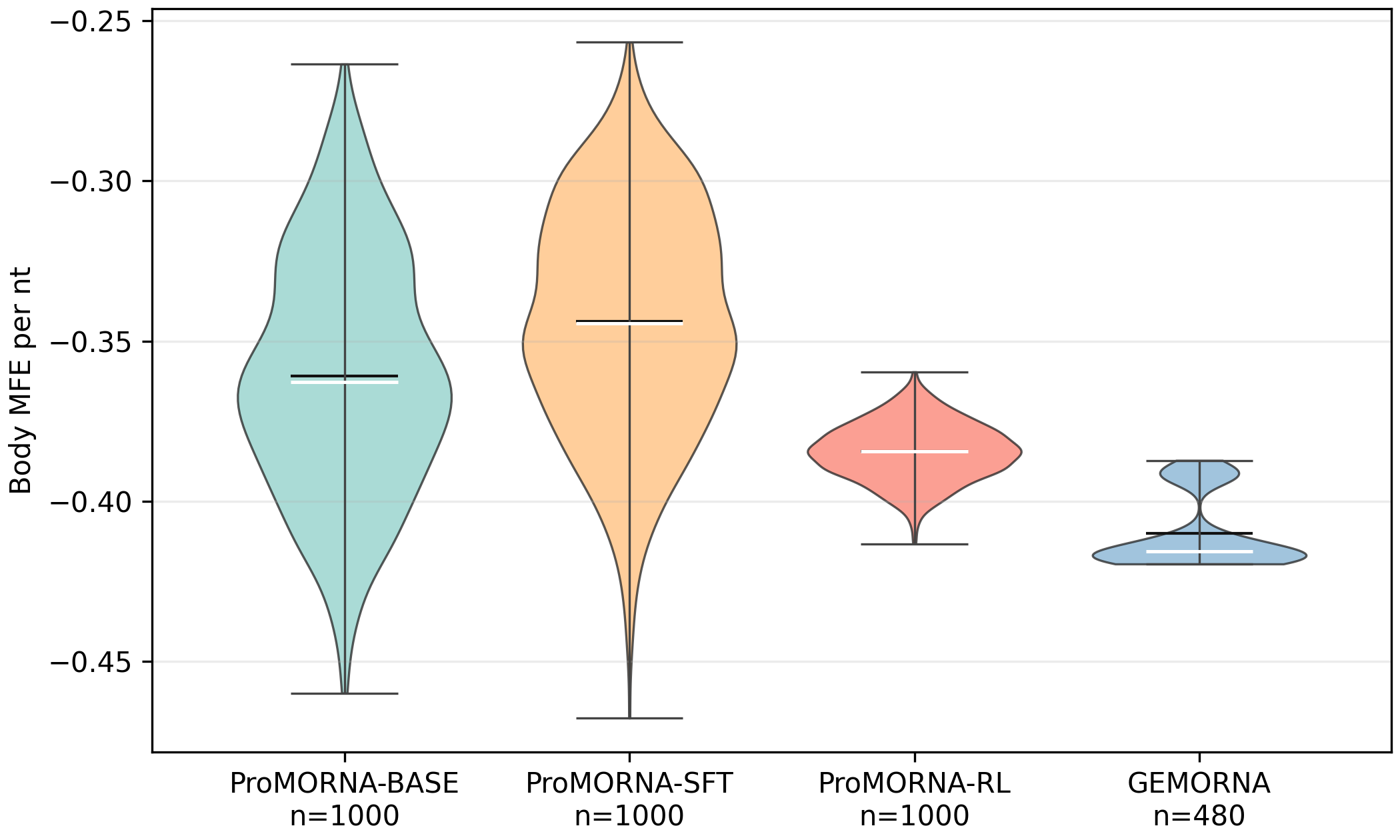}
    \caption{Body MFE}
    \label{fig:seq_bmfe}
  \end{subfigure}\hfill
  \begin{subfigure}[b]{0.3\linewidth}
    \centering
    \includegraphics[width=\linewidth]{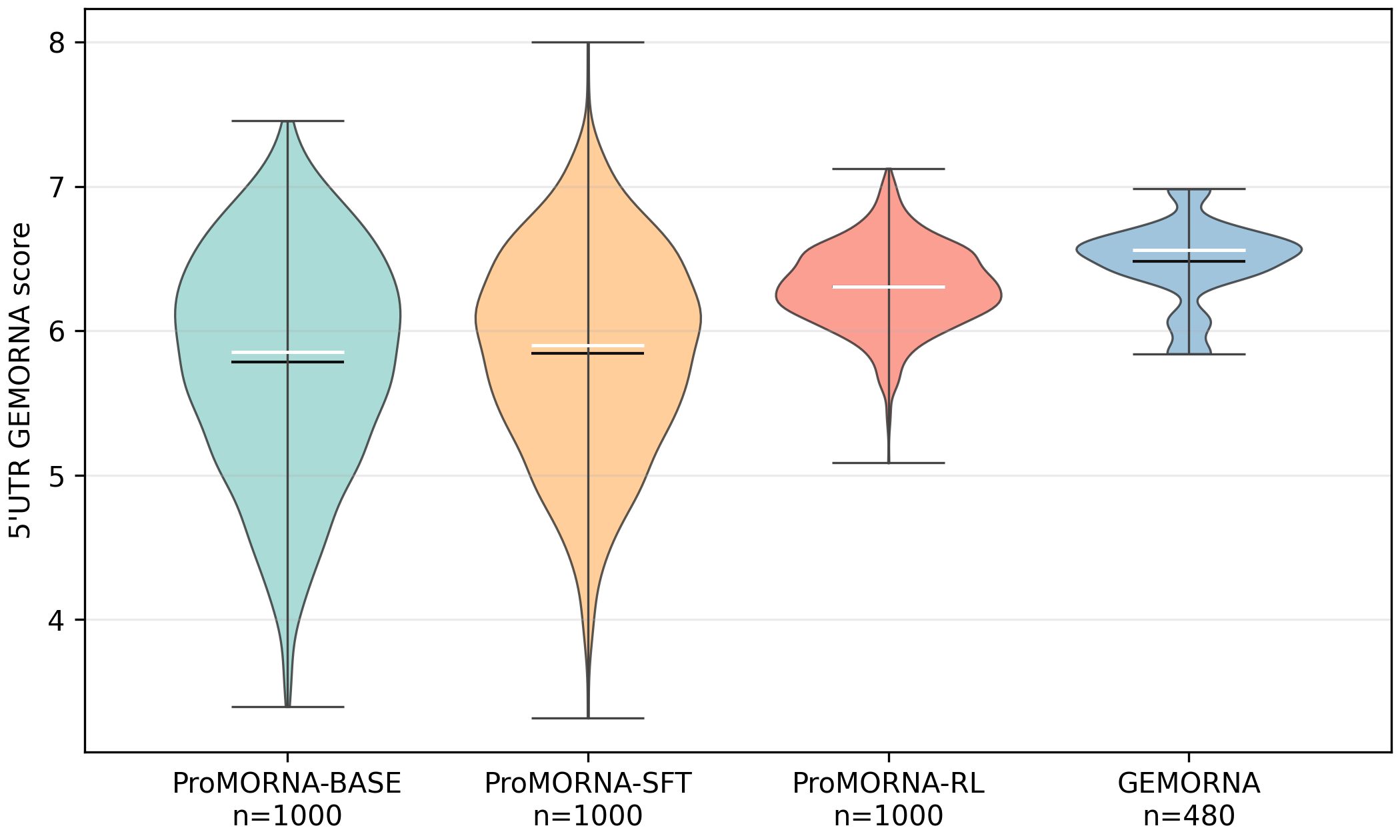}
    \caption{5' UTR GEMORNA Score}
    \label{fig:seq_5u_gmr}
  \end{subfigure}\hfill
  \begin{subfigure}[b]{0.3\linewidth}
    \centering
    \includegraphics[width=\linewidth]{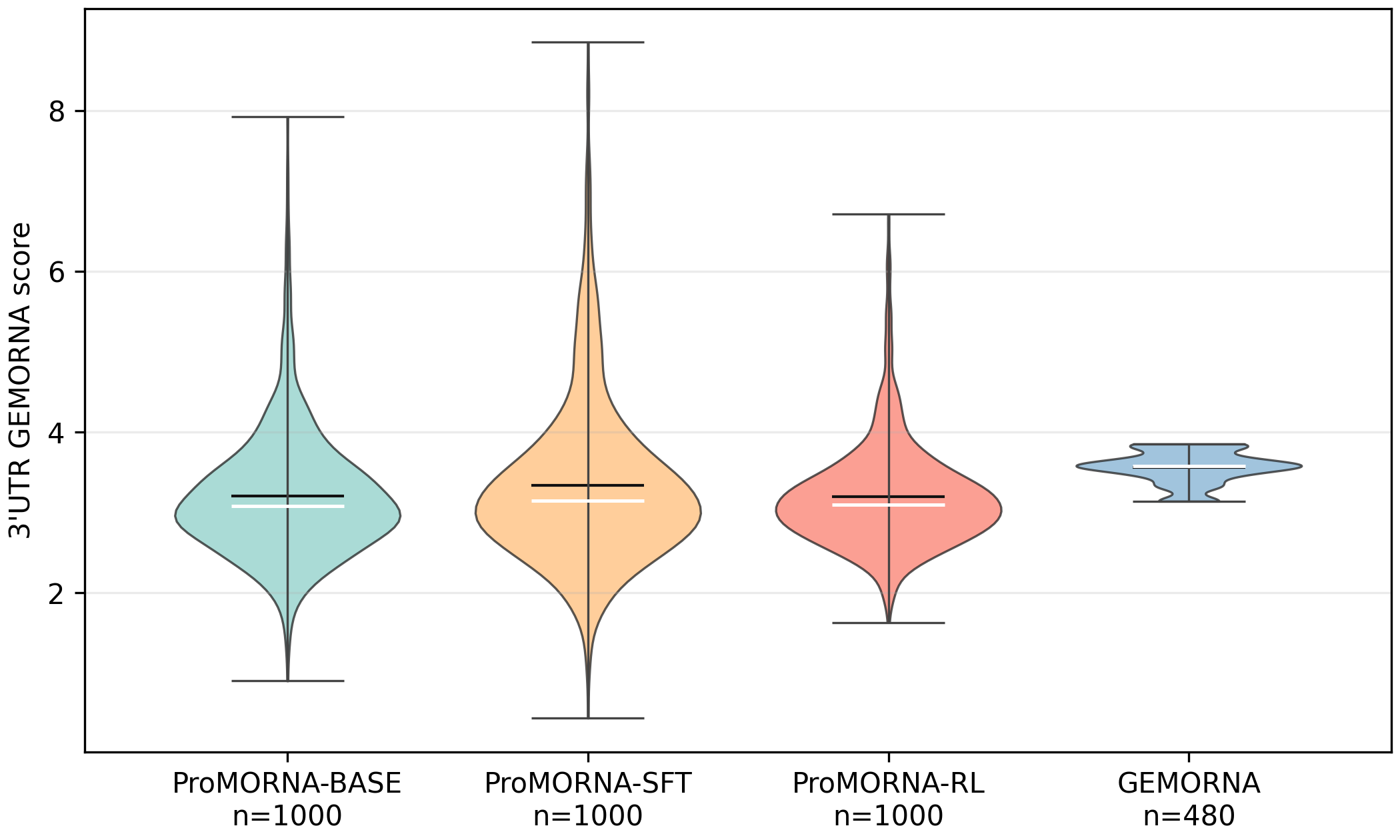}
    \caption{3' UTR GEMORNA Score}
    \label{fig:seq_3u_gmr}
  \end{subfigure}

  \vspace{0.8em}

  \begin{subfigure}[b]{0.3\linewidth}
    \centering
    \hspace*{\fill}
    \includegraphics[width=\linewidth]{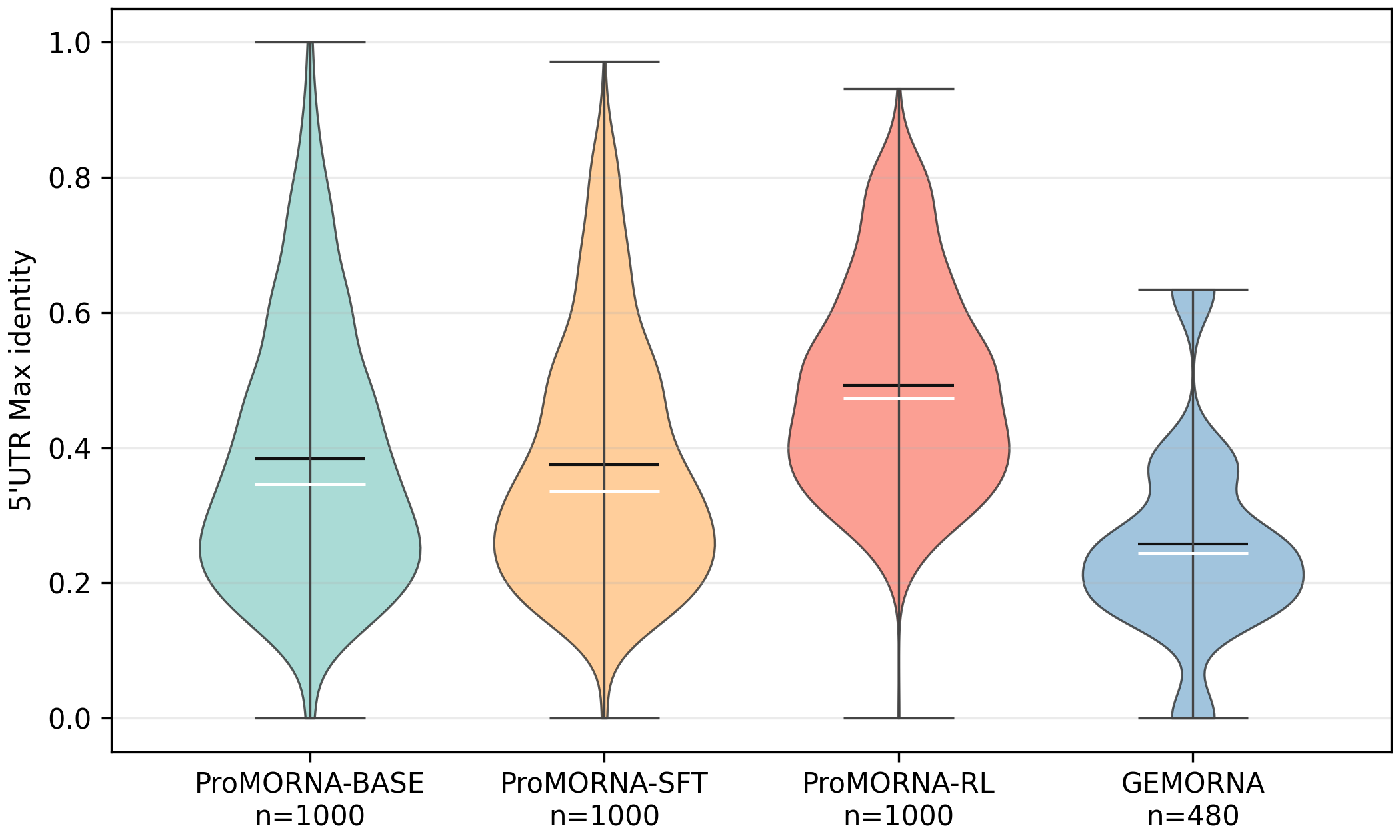}
    \caption{5' UTR Max Identity}
    \label{fig:seq_5u_mis}
  \end{subfigure}\hspace{0.05\linewidth}
  \begin{subfigure}[b]{0.3\linewidth}
    \centering
    \includegraphics[width=\linewidth]{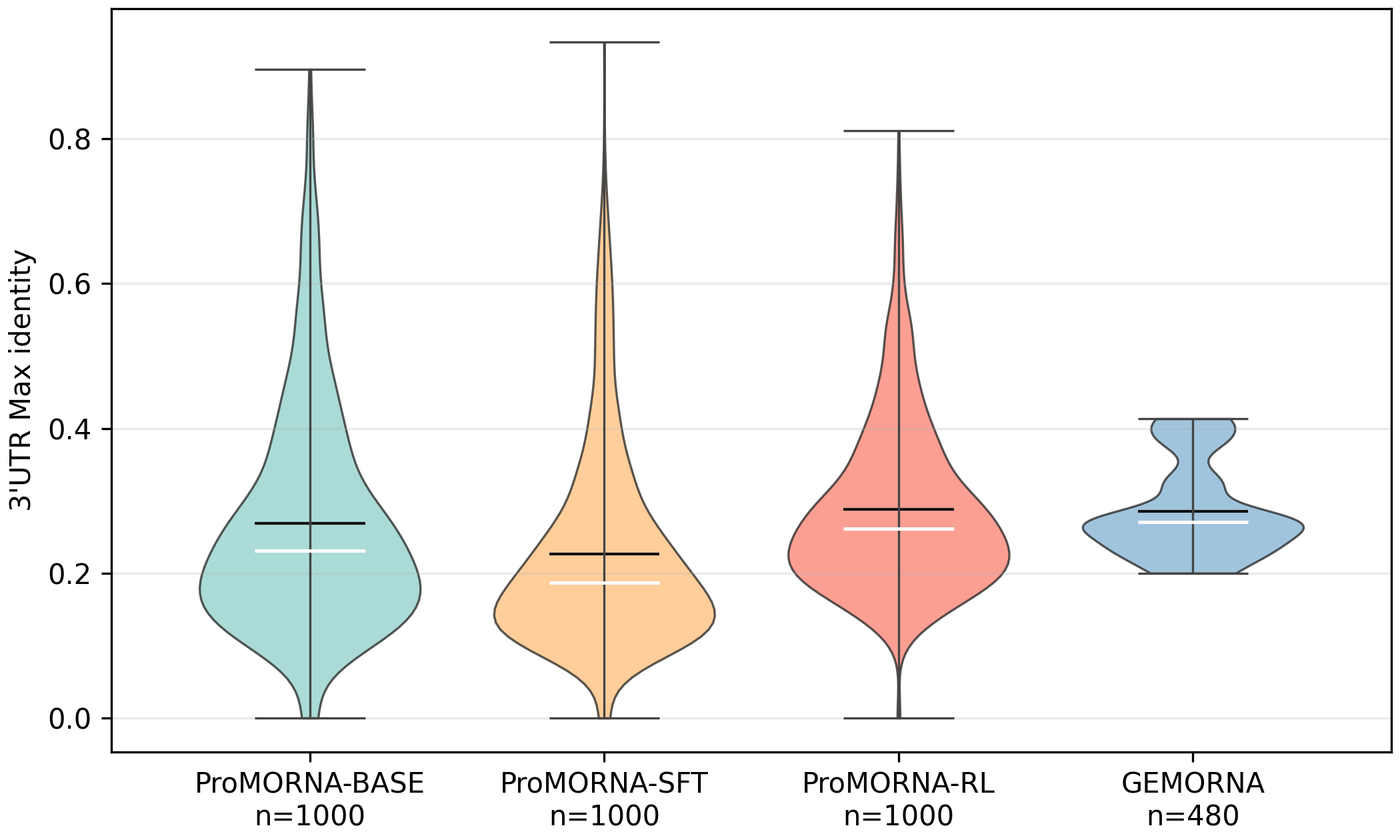}
    \caption{3' UTR Max Identity}
    \label{fig:seq_3u_mis}
  \end{subfigure}

  \caption{
  Additional sequence-property analysis of valid, unique firefly-luciferase mRNA candidates. The GEMORNA UTR scores are computed using the external UTR scoring models from GEMORNA~\cite{zhang2025deep}.
  }
  \label{fig:seqprop}
\end{figure}

To better understand the sequence-level mechanics driving these results, \cref{fig:seqprop} details additional sequence properties. Although not primary reward objectives, these metrics provide valuable mechanistic insights into how ProMORNA shapes nucleotide and codon usage and alters regional mRNA secondary structure.

As illustrated in \cref{fig:seq_gc}, ProMORNA-RL produces a moderate increase in median GC content relative to both the BASE and SFT models, accompanied by a dramatic reduction in variance. While higher GC content is linked to increased mRNA abundance in mammalian cells~\cite{kudla2006high}, excessive GC enrichment can deleteriously impact local secondary structure, perturb optimal codon context, and hinder downstream sequence manufacturability. ProMORNA's moderate increase aligns with a balanced optimization strategy, avoiding the extremes of a single-proxy focus.

A similar multi-objective trade-off is observed in the Codon Adaptation Index (CAI)~\cite{sharp1987codon} (\cref{fig:seq_cai}). While GEMORNA exhibits near-maximal CAI values due to its strong preference for high-frequency synonymous codons, ProMORNA-RL converges to a tightly concentrated, moderate CAI regime around $0.78$, which is slightly lower than those of the BASE and SFT models. Because CAI is not explicitly included in our reward formulation, this distribution indicates that RL post-training actively trades off greedy codon optimality to satisfy competing global constraints, such as minimizing U-content, enhancing full-length thermodynamic stability, and maintaining favorable local folding topologies~\cite{mauger2019mrna,metkar2024tailor}. Notably, despite having a lower CAI than GEMORNA, ProMORNA-RL achieves superior predicted translation rates under the independent Saluki evaluator and strong \textit{in vitro} reporter activity, demonstrating that maximizing heuristic codon usage alone is neither necessary nor sufficient for optimal protein expression.

\Cref{fig:seq_lmfe,fig:seq_bmfe} map region-specific mRNA secondary structure by assessing normalized MFE of different regions. Biologically, a higher MFE in the 5' leader region (the 5' UTR and first 10 CDS nucleotides) indicates looser secondary structure, which is known to facilitate efficient ribosome scanning~\cite{tuller2015multiple,mauger2019mrna}. Conversely, tighter secondary structures in the remainder of the mRNA body (the rest of the CDS and the 3' UTR) broadly enhance transcript stability and protein output, corresponding to a lower MFE~\cite{leppek2022combinatorial}. Compared to the BASE and SFT models, ProMORNA-RL closely aligns with these biological design principles: it shifts the 5' leader MFE upward to minimize translation initiation barriers, while simultaneously driving the body MFE downward to stabilize the rest of the transcript. Although GEMORNA achieves more extreme MFE values in both regions, likely due to its reliance on a heavily curated sequence pool, ProMORNA autonomously learns to optimize these regional folding topologies while retaining structural diversity.

\Cref{fig:seq_5u_gmr,fig:seq_3u_gmr} evaluate the generated sequences using external GEMORNA UTR scoring models, which estimate UTR quality based on predictors trained for high Mean Ribosome Load in 5' UTRs and high stability in 3' UTRs~\cite{zhang2025deep}.
While GEMORNA exhibits concentrated, high-scoring distributions due to sampling UTRs from its own pre-optimized pool, ProMORNA-RL achieves a highly competitive 5' UTR score distribution. Notably, it demonstrates significant improvement and reduced variance relative to the BASE and SFT models, all while maintaining broader sequence diversity.

We also assessed the sequence similarities between generated and natural UTRs using BLAST~\cite{altschul1990basic}. A low max identity score (MIS) indicates minimal homology between the AI-generated UTRs and naturally occurring UTRs in the reference database. As shown in \cref{fig:seq_5u_mis,fig:seq_3u_mis}, the MIS distributions for the 5' and 3' UTRs reveal that the ProMORNA models generate sequences with low to moderate homology to known natural transcripts. This confirms that the pipeline performs genuine \textit{de novo} generation rather than merely memorizing sequences from reference databases.

Furthermore, ProMORNA-RL's two regulatory regions display distinct sequence-space dynamics.
For the 5' UTR (\cref{fig:seq_5u_mis}), ProMORNA-RL demonstrates a moderate upward shift in median MIS compared to the BASE and SFT models; for the 3' UTR (\cref{fig:seq_3u_mis}), it maintains a persistently low sequence similarity. Taken together, these regional differences may highlight a sophisticated, region-specific optimization strategy within ProMORNA-RL: it autonomously converges toward natural-like motifs where translational initiation constraints are strict (5' UTR), while broadly exploring entirely novel, artificial sequence spaces where structural flexibility permits (3' UTR).

\end{appendices}

\end{document}